\documentclass[journal]{IEEEtran}
\usepackage{amsmath,amssymb,amsfonts}
\usepackage{array}
\usepackage{etoolbox}
\usepackage{textcomp}
\usepackage{url}
\usepackage{verbatim}
\usepackage{graphicx}
\usepackage{tabularx}
\usepackage{amsthm}%
\usepackage{mathrsfs}%
\usepackage{xcolor}%
\usepackage{multirow}
\usepackage{siunitx}
\usepackage{manyfoot}%
\usepackage{booktabs}%
\usepackage{hhline}
\usepackage{listings}%
\usepackage[linesnumbered, ruled, vlined]{algorithm2e}
\usepackage{tikz}
\usepackage{cite}
\usepackage{makecell} 
\usepackage{hyperref}
\usepackage{soul} 
\usepackage{subcaption}

\makeatletter
\patchcmd{\@makecaption}{\scshape}{}{}{}
\makeatother
\def\tablename{Table}

\def\BibTeX{{\rm B\kern-.05em{\sc i\kern-.025em b}\kern-.08em
    T\kern-.1667em\lower.7ex\hbox{E}\kern-.125emX}}

\begin{document}
\bstctlcite{IEEEexample:BSTcontrol}

\title{Dynamic Buffers: Cost-Efficient Planning for Tabletop Rearrangement with Stacking}

\author{
    Arman Barghi,
    Hamed Hosseini,
    Seraj Ghasemi,
    Mehdi Tale Masouleh,
    Ahmad Kalhor%
    \thanks{All authors are with the Human and Robot Interaction Laboratory, School of Electrical and Computer Engineering, University of Tehran, Tehran, Iran (e-mail: arman.barghi@ut.ac.ir, hosseini.hamed@ut.ac.ir, seraj.ghasemi@ut.ac.ir, m.t.masouleh@ut.ac.ir, akalhor@ut.ac.ir).}
    \thanks{%
    \copyright\ 2026 IEEE. Personal use of this material is permitted. Permission from IEEE must be obtained for all other uses, in any current or future media, including reprinting/republishing this material for advertising or promotional purposes, creating new collective works, for resale or redistribution to servers or lists, or reuse of any copyrighted component of this work in other works.
    \par This is the accepted manuscript of A. Barghi, H. Hosseini, S. Ghasemi, M. Tale Masouleh, and A. Kalhor, ``Dynamic Buffers: Cost-Efficient Planning for Tabletop Rearrangement with Stacking,'' \emph{IEEE Transactions on Automation Science and Engineering}, 2026, doi: 10.1109/TASE.2026.3735598. IEEE Xplore record: \url{https://ieeexplore.ieee.org/document/11701320}.}%
}

\maketitle

\begin{abstract}
Rearranging objects in cluttered tabletop environments remains a long-standing challenge in robotics. Classical planners often generate inefficient, high-cost plans by moving objects individually and using fixed buffers---temporary spaces such as unoccupied tabletop regions or static stacks---to resolve conflicts. When only free tabletop locations are used as buffers, planning in dense scenes becomes difficult, since placing an object can prevent others from reaching their goals and complicate subsequent planning. Stacking provides additional buffer capacity, but conventional stacking is static: the base object cannot be relocated while holding another, which limits efficiency. To address these limitations, a novel planning primitive called the \emph{Dynamic Buffer} is introduced. Inspired by human grouping strategies, it enables robots to form temporary, movable stacks that can be transported as a unit. Dynamic buffers improve feasibility and efficiency in dense layouts and reduce travel in large-scale scenarios where buffer space is ample. Compared with a state-of-the-art rearrangement planner, the proposed approach reduces total manipulator travel cost by 13.48\% in dense stationary-manipulator scenarios and by 4.34\% in large-scale, low-density scenarios involving mobile manipulators. Practicality is validated through execution on a Delta robot with a parallel-jaw gripper. These results establish dynamic buffering as a key primitive for cost-efficient and robust rearrangement planning.
\end{abstract}


\section{Introduction}\label{sec:introduction}
The growing field of Embodied AI is focused on creating autonomous systems that can physically interact with and modify their environments to achieve goals. A crucial aspect of this interaction is the ability to rearrange objects, a task identified as a canonical challenge for embodied agents~\cite{batra2020embodied}. This capability is fundamental to a wide range of applications, from industrial automation, such as sorting and packing on a production line~\cite{labbe2020mcts}, to domestic services like tidying a room~\cite{wu2023tidybot}. The comprehensive nature of rearrangement, which unifies perception, planning, and manipulation, makes it a robust benchmark for progress in autonomous robotics.
\begin{figure}[t]
	\centerline{\includegraphics[ width = 0.47\textwidth]{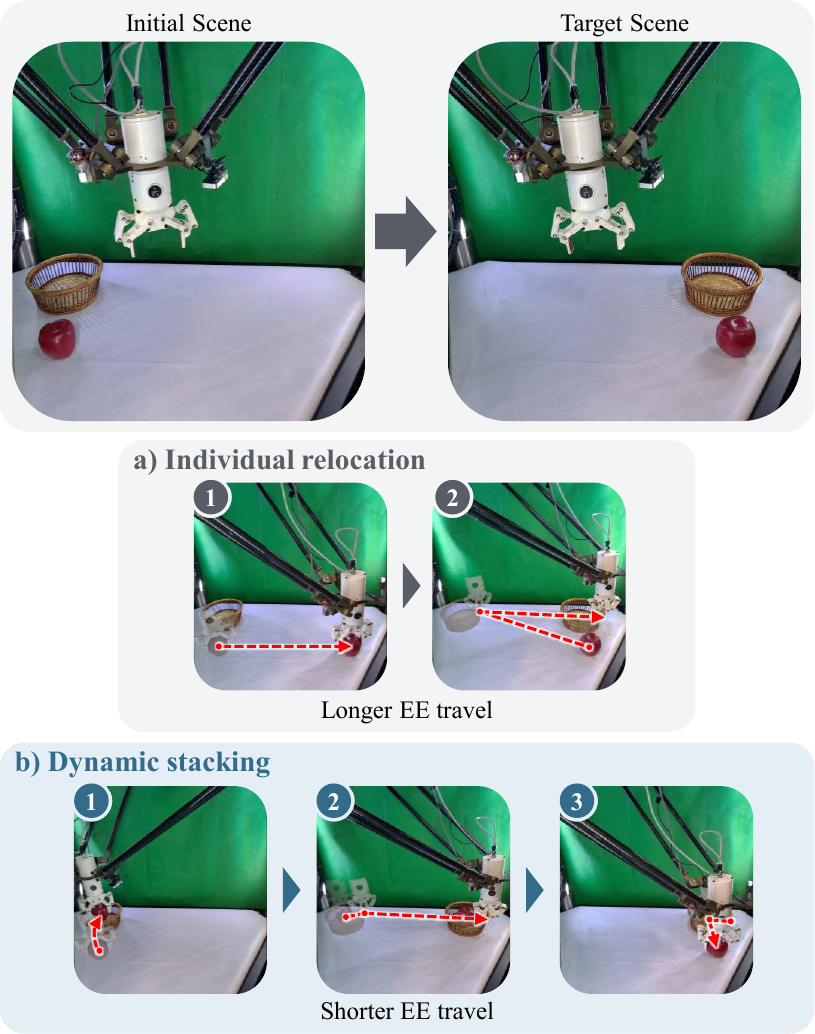}}
	\caption{Illustration of two rearrangement plans for a stationary Delta robot. (a) Individual relocation without stacking requires separate object movements and results in longer end-effector (EE) travel. (b) The proposed Dynamic Buffer plan first forms a temporary stack and transports it as a unit, reducing EE travel despite the additional stacking action.}
	\label{fig:robot}
\end{figure}

\begin{figure*}[t]
	\centerline{\includegraphics[ width = 0.99\textwidth]{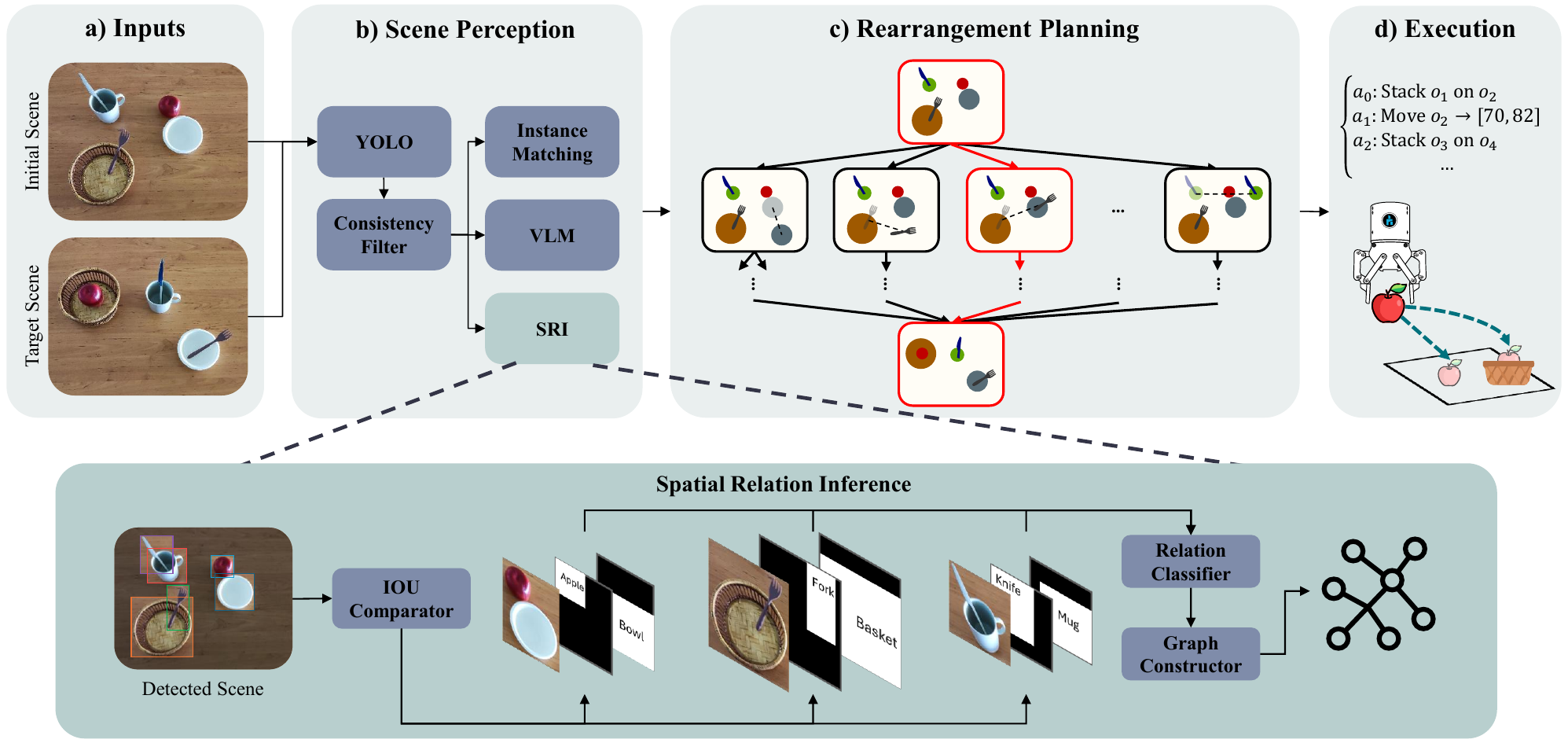}}
	\caption{Overall framework of the proposed robotic rearrangement system, illustrating the multi-stage pipeline from initial and target scene perception to rearrangement planning and execution.}
	\label{fig:graphical_abstract}
\end{figure*}

Rearrangement planning---a fundamental challenge within task and motion planning (TAMP)~\cite{garrett2021tamp}---aims to find a sequence of actions to move a set of objects from an initial arrangement to a target configuration~\cite{labbe2020mcts}. This process requires a robotic system to first perceive the environment to establish these configurations. Subsequently, the planner must address the core challenge of objects obstructing each other's target locations. It computes a sequence of pick-and-place actions, often using temporary positions, to resolve these conflicts and ensure that each object can be moved to its destination without collision. Once a complete plan is generated, it is passed to the robot for execution. The overall framework, from visual input to robot action, is illustrated in Fig.~\ref{fig:graphical_abstract}.

A key formulation of this challenge is the Tabletop Object Rearrangement with Overhand Grasps (TORO) problem, which seeks a plan that minimizes a specific cost metric, primarily by reducing the number of pick-and-place actions~\cite{han2018toro}. 
The problem is fundamentally NP-hard, with its complexity stemming from object dependencies that arise when objects obstruct each other's goal positions~\cite{han2018toro}. These dependencies are typically broken by using ``buffers''---temporary locations for obstructing objects. While finding a feasible plan is difficult, optimizing for cost-efficiency is even harder, as the selection of buffer locations directly impacts the total manipulator travel distance~\cite{gao2022trlb, hu2025strap}. 
Traditional approaches often struggle in high-density scenarios where limited buffer space is available, leading to limitations in both feasibility and plan quality.

In high-density scenarios, stacking objects can create valuable buffer space. While prior works utilize stacking for retrieval path-clearing~\cite{lee2022retrieval}, rearrangement buffering~\cite{gao2025orla}, or shelf layouts~\cite{chen2024osamipstack}, they uniformly treat stacks as static, locking the base object in place. This static approach is inefficient compared to human strategies, where temporary groups (e.g., stacked bowls) are transported in a single action. Inspired by these human strategies, this paper introduces a novel planning primitive termed the ``Dynamic Buffer''. By purposefully stacking objects and relocating the entire unit via its base, multiple single-object actions are consolidated into one efficient multi-object move, significantly reducing travel distance (Fig.~\ref{fig:robot}).
To the authors' knowledge, this is the first work to treat stacks not merely as static buffers, but as dynamic, transportable units to improve cost-efficiency in a long-horizon rearrangement plan.

The principal contributions of this paper are twofold:
\begin{itemize}
\item The \textit{Dynamic Buffer}, a novel planning primitive that enables efficient multi-object manipulation by treating temporary stacks as transportable units, fundamentally shifting stacking from a static tool to a dynamic one to reduce manipulator travel distance.
\item An integrated, stack-aware rearrangement system featuring a perception module for identifying stable stacking relationships, which enables a search-based planner to effectively utilize dynamic buffers.
\end{itemize}

\begin{table*}[tp]
	\centering
	\caption{Comparison of robotic object rearrangement planning methods, highlighting buffering strategies, object-as-buffer usage (SS: Static Stacking, DS: Dynamic Stacking), placement decision logic, optimization objectives, and manipulator applicability.}
	\label{tab:literature_compare}
	\begin{tabular*}{\textwidth}{@{\extracolsep{\fill}}
			p{0.135\textwidth} 
			p{0.225\textwidth} 
			p{0.16\textwidth} 
			p{0.145\textwidth} 
			p{0.18\textwidth} c @{}}
		\toprule
		\textbf{Paper} &
		\textbf{\makecell{Buffering\\Method}} &
		\textbf{\makecell{Unoccupied Object\\Goal Strategy}} &
		\textbf{\makecell{Occupied Object\\Goal Strategy}} &
		\textbf{\makecell{Optimization\\Objective}} &
		\textbf{\makecell{Mobile\\Manip.}} \\
		\midrule
		TORO~\cite{han2018toro} (2018) & External & Direct to Goal & To External Buffer & Cost-Eff. (Manip. Travel) & $\times$ \\
		MCTS~\cite{labbe2020mcts} (2020) & Internal (Unoccupied Space) & Direct to Goal & Move Obstructing Obj. & Feasibility & $\times$ \\
		TRLB~\cite{gao2022trlb} (2022) & Internal (Unoccupied Space) & Direct to Goal & To Lazy Buffer & Cost-Eff. (Num. of Actions) & $\times$ \\
		HetGNN~\cite{lou2023hetgnn} (2023) & Internal (Unoccupied Space) & Policy-Learned & Policy-Learned Action & Cost-Eff. (Action Type) & $\times$ \\
		PMMR~\cite{huang2024pmmr} (2024) & Internal (Unoccupied Space) & Prioritized Direct to Goal & To Sampled Buffer & Cost-Eff. (Obj. Travel) & $\times$ \\
		ORLA*~\cite{gao2025orla} (2025) & Internal (Unoccupied Space + SS) & Direct to Goal & To Lazy Buffer & Cost-Eff. (Manip. Travel) & \checkmark \\
		STRAP~\cite{hu2025strap} (2025) & Internal (Unoccupied Space) & Direct to Goal & To Sampled Buffer & Cost-Eff. (Manip. Travel) & \checkmark \\
		\textbf{This Work} & \textbf{Internal (Unoccupied Space + DS)} & \textbf{Prioritized Stacking} & \textbf{To Sampled Buffer} & \textbf{Cost-Eff. (Manip. Travel)} & \textbf{\checkmark} \\
		\bottomrule
	\end{tabular*}
\end{table*}

The remainder of this paper is structured as follows. Section~\ref{sec:related-work} provides a comprehensive overview of related work in robotic object perception, rearrangement planning, and multi-object manipulation, followed by Section~\ref{sec:formulation}, which presents the problem formulation. Section~\ref{sec:method} details the proposed methodology, including the scene perception, search algorithm, expansion strategy, and plan refinement. Experimental validation and a thorough analysis of results are presented in Section~\ref{sec:results}, followed by concluding remarks and future directions in Section~\ref{sec:conclusion}.
\section{Related Work}\label{sec:related-work}
This section surveys literature on robotic object rearrangement, proceeding from scene perception to planning strategies and multi-object manipulation. In doing so, it contextualizes existing buffering approaches and highlights the need for dynamic, transportable stacks.

\subsection{Scene Perception}
A fundamental challenge in rearrangement planning is perceiving the environment to define the goal and represent the scene. The goal itself can be specified in different ways---as an explicit set of object poses~\cite{gao2022trlb, gao2025orla, hu2025strap}, an implicit objective such as a “tidiness score”~\cite{kee2025tsmcts}, or a natural language command~\cite{chang2024lgmcts}. In some cases, the goal must instead be generated from commonsense rules~\cite{zhai2024sgbot}.
To reason about these environments, planners commonly map the perceived data into graph-based representations. These graphs efficiently capture relational semantics through hierarchical or heterogeneous structures~\cite{zhu2021symbolic, lou2023hetgnn} and can explicitly encode the rearrangement task by representing corresponding start-goal object pairs as unified nodes~\cite{qureshi2021nerp}.
Additionally, visually defined goals~\cite{labbe2020mcts, huang2024pmmr} introduce the object-matching problem---determining correspondences between initial and target scenes. While these correspondences are often assumed to be known~\cite{gao2022trlb, gao2025orla, hu2025strap}, integrating matching into the perception pipeline provides a more realistic unification of scene understanding and goal inference~\cite{qureshi2021nerp, lou2023hetgnn, xu2024see}.

\subsection{Rearrangement Planning}
Rearrangement planning is typically categorized into retrieval-based tasks, targeting a single object~\cite{cheong2020relocate, kang2023retrieval, ren2024retrieval}, and goal-oriented tasks, where all objects have designated target configurations. While goal-oriented problems span diverse domains ranging from room-scale rearrangement~\cite{batra2020embodied, wang2020scenemover, mirakhor2024multiroom} to confined spaces~\cite{ren2024msmcts}, this work focuses on tabletop manipulation, where resolving object dependencies in dense clutter remains the primary challenge~\cite{labbe2020mcts}. 

A key design choice for tabletop planners is the buffering strategy employed to resolve these dependencies. Approaches range from utilizing dedicated off-table \textit{external} areas~\cite{han2018toro, gao2023runningbuf} to more challenging on-table \textit{internal} spaces that risk creating new blockages~\cite{gao2022trlb}. Recently, objects themselves have been leveraged as temporary buffers for other objects in dense scenes~\cite{gao2025orla}. Table~\ref{tab:literature_compare} contextualizes the proposed method by comparing prominent approaches across these design axes. 

A planner’s placement strategy dictates buffer utilization. For occupied targets, tactics include moving the obstructing object~\cite{labbe2020mcts}, relocating the current object to a pre-defined buffer~\cite{han2018toro, hu2025strap}, or assigning buffers lazily~\cite{gao2022trlb, gao2025orla}. Unoccupied targets typically prompt direct moves, though alternative placements can optimize long-term sequences~\cite{huang2024pmmr}. In contrast, the proposed strategy can prioritize stacking even when a goal is unoccupied, enabling subsequent, efficient multi-object moves.

Rearrangement planning approaches are distinguished by their objective. While some methods prioritize finding a feasible solution without an explicit cost function~\cite{labbe2020mcts, qureshi2021nerp}, cost-efficient approaches aim to generate high-quality plans. However, the notion of ``high-quality'' is defined by various optimization metrics. These include minimizing the number of actions~\cite{gao2022trlb}, the number of running buffers~\cite{gao2023runningbuf}, or costs based on action type (e.g., prioritizing pushing over picking)~\cite{lou2023hetgnn}. Other works focus on path-based metrics, such as the total distance traveled by objects~\cite{huang2024pmmr} or the manipulator~\cite{han2018toro,gao2025orla,hu2025strap}. The latter is a highly practical objective, as it directly correlates with execution time and energy consumption.
Moreover, to scale throughput and workspace size, recent works employ multi-arm systems~\cite{gao2024holistic} or mobile manipulators~\cite{gao2025orla, hu2025strap}, where minimizing base travel distance becomes the primary focus.

Algorithmically, rearrangement planning relies on search-based and learning-based paradigms. Approaches such as Monte Carlo Tree Search (MCTS)~\cite{labbe2020mcts} and certain learning models~\cite{qureshi2021nerp} prioritize feasibility over explicit cost optimization. While some learning approaches optimize travel distance, they generally avoid the high geometric dependencies of dense tabletops, focusing instead on single-target retrieval~\cite{ren2024retrieval} or room-scale tasks with decoupled, reactive buffering~\cite{mirakhor2024multiroom}. 
The combinatorial explosion of exact placement and sequencing in dense environments severely challenges learning models, forcing simplifications like abstracting environments into rigid grids with interchangeable test tubes~\cite{chen2026testtube} or restricting actions to directional pushes~\cite{wang2020scenemover}. Consequently, minimizing travel distance in full-scene tabletop rearrangement requires exact geometric reasoning and integrated buffer allocation, typically achieved through heuristic search methods such as $A^*$~\cite{gao2025orla, hu2025strap}.

\subsection{Multi-Object Manipulation}
The choice of manipulation primitive is critical for rearrangement, spanning non-prehensile sorting~\cite{song2020sort}, precise prehensile grasping~\cite{gao2022trlb, gao2025orla}, and hybrid methods that combine both~\cite{huang2024pmmr}. Scaling up efficiency, however, requires moving beyond a one-object-at-a-time paradigm. 
Research in multi-object manipulation addresses this at varying levels of complexity. At its most direct, a robot can plan to grasp several already-adjacent objects simultaneously~\cite{agboh2023multiobj}. When objects are not yet grouped, the ``Busboy Problem'' formulation requires the robot to first execute setup actions---stacking one item into another (``stack-grasp'')---to create a compact, graspable configuration before lifting them as a unit~\cite{srinivas2023multiobj}. A third approach adds semantic understanding, enabling the robot to detect existing stable stacks in the environment and decide whether moving an entire stack is more efficient than handling items individually~\cite{wu2023multiobj}.

While foundational works use stacking for aggregation or removal, recent dense rearrangement methods utilize it for buffering but treat stacks as \textit{static}, immobilizing the base object~\cite{gao2025orla}. Addressing this limitation, the proposed Dynamic Buffer primitive enables active creation and relocation of temporary, transportable stacks within the workspace to strategically free space, resulting in more flexible and cost-efficient planning. To demonstrate its effectiveness, this primitive is integrated into the state-of-the-art $A^*$-based STRAP planner~\cite{hu2025strap} and a feasibility-focused MCTS planner~\cite{labbe2020mcts}. The approach is evaluated across stationary and mobile manipulators, benchmarking against the static stacking strategy from \cite{gao2025orla}.

\section{Problem Statement}\label{sec:formulation}
The robotic object rearrangement problem seeks to transform an initial configuration into a target configuration. The task begins by processing visual observations of the initial and target scenes to extract structured representations of the objects, their positions, and pairwise relations. Establishing consistent correspondences across scenes yields a unified set of objects \( \mathcal{V} = \{o_1, \dots, o_n\} \), which forms the basis of the state representation.

\begin{figure}[t]
	\centerline{\includegraphics[ width = 0.5\textwidth]{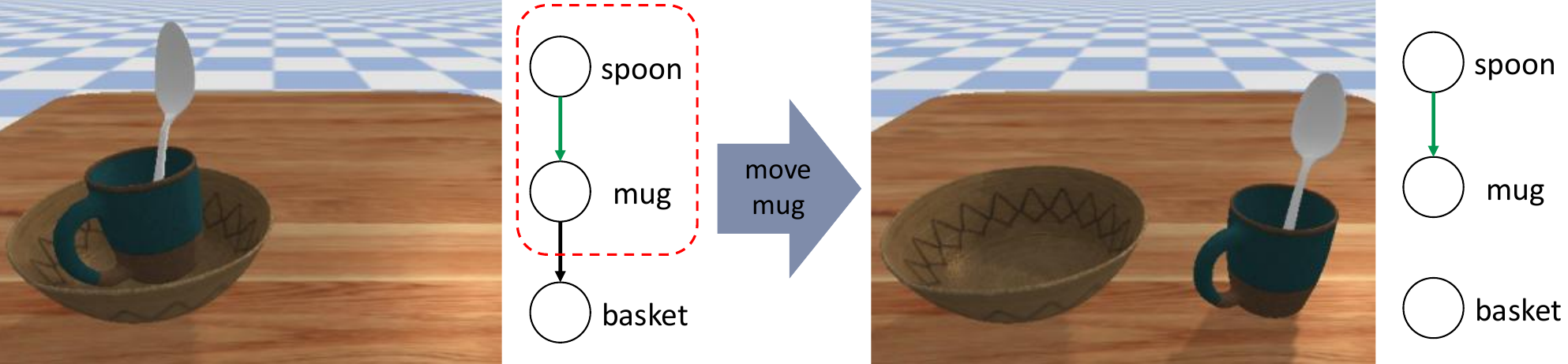}}
	\caption{The Dynamic Buffer concept in action: when a base object (mug) is moved, the stacked object (spoon) moves along with it, maintaining their relative stacking relationship, as also shown in the accompanying relational graphs.}
	\label{fig:db_example}
\end{figure}

A state is defined as a valid and collision-free configuration of the environment, represented as \( s = (\mathcal{G}, p_m) \), where \( \mathcal{G} = (\mathcal{V}, \mathcal{E}) \) is a relational graph encoding the object arrangement and \( p_m \) denotes the manipulator's position. Each node \( o_i \in \mathcal{V} \) encodes its current position \( p_i \) and relevant object attributes. The edges \( \mathcal{E} \) specify stacking relationships, with an edge \( e_{ij} \) indicating that object \( o_i \) is stacked on object \( o_j \). The manipulator starts at a predefined home position \( p_m^{(0)} \) and returns to it upon task completion.
The goal state \( s^{(T)} \) specifies the desired configuration of the objects. For each object, this goal can be defined either by an absolute target position or by a relative stacking relation with another object. The robotic rearrangement problem is therefore to compute a collision-free and cost-efficient plan that transforms the environment from the initial state \( s^{(0)} \) to the target state \( s^{(T)} \).

Executing a valid action \( a \in \mathcal{A} \) from state \( s \) transitions the system to a new state \( s' = \mathcal{T}(s, a) \), updates the manipulator's position \( p_m \), and incurs a cost \( c(a) \).
In this framework, two action primitives can be performed on an object $o_i$:
\begin{itemize}
	\item \textbf{Move}: Relocate $o_i$ from its current position $p_i$ to an unoccupied position $p_i'$ on the workspace.
	\item \textbf{Stack}: Place $o_i$ from its current position $p_i$ into a valid, unoccupied base $o_j$ at position $p_j$, establishing a stacking relationship (represented by updating the edge $e_{ij} = 1$).
\end{itemize}
A core feature of this approach is the concept of a Dynamic Buffer: when an object \( o_j \) is moved to a new position, all objects stacked upon it (i.e., with an edge \( e_{ij} = 1 \), and their transitive dependents) move accordingly, maintaining their relative stacking relationships. This dynamic transport, such as moving a mug containing a spoon (Fig.~\ref{fig:db_example}), enables highly efficient rearrangement.

\begin{figure}[t!]
    \centering
    
    \begin{subfigure}[t]{0.24\textwidth}
        \centering
        \includegraphics[width=\linewidth]{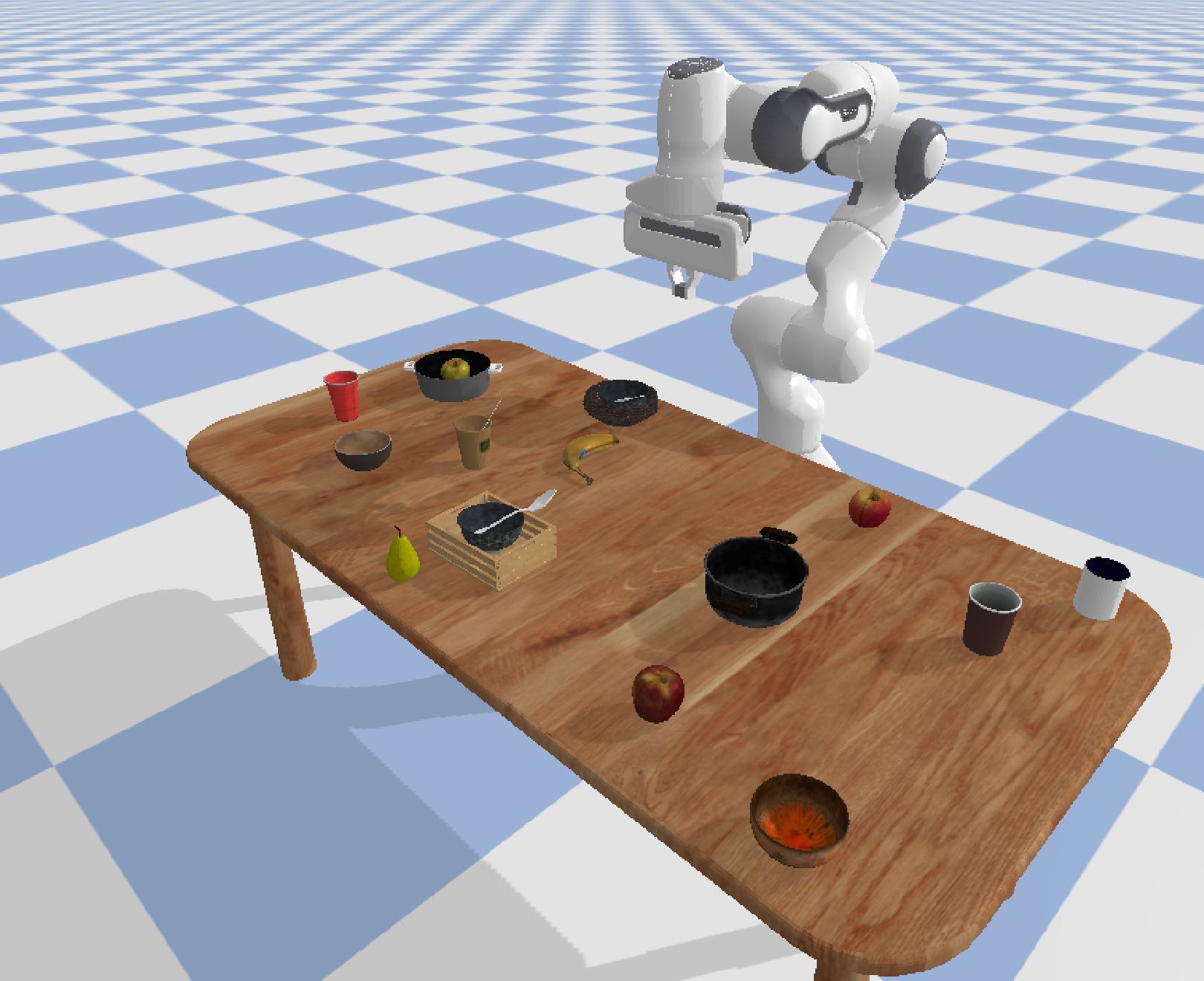}
    \end{subfigure}
    \hfill
    \begin{subfigure}[t]{0.24\textwidth}
        \centering
        \includegraphics[width=\linewidth]{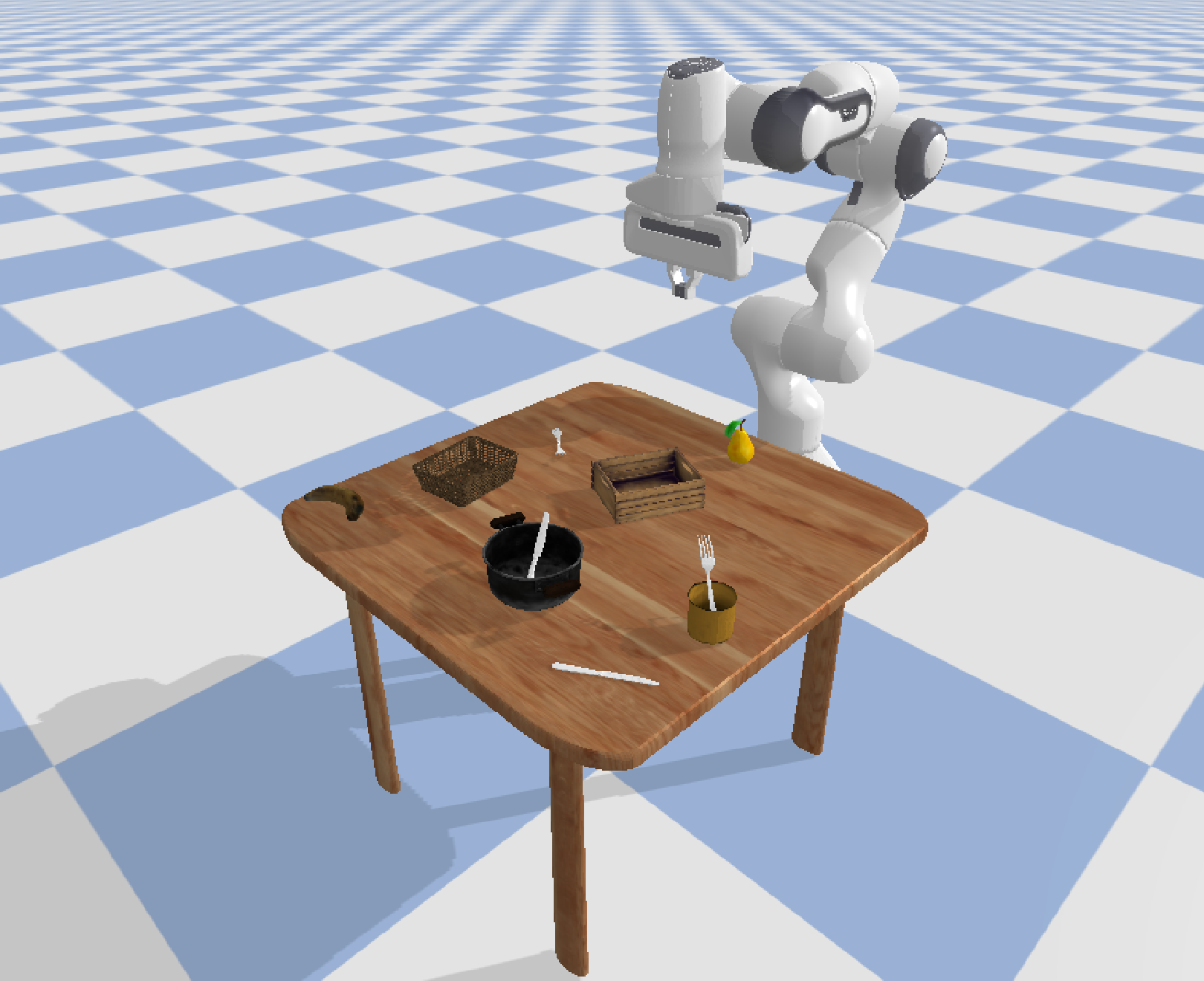}
	\end{subfigure}

	\caption{The two manipulator configurations used in the simulation: (Left) mobile manipulator for larger tables requiring base movement, and (Right) stationary manipulator for smaller workspaces where the base remains fixed.}
    \label{fig:sim_env}
\end{figure}

\begin{figure}[t]
	\centerline{\includegraphics[ width = 0.4\textwidth]{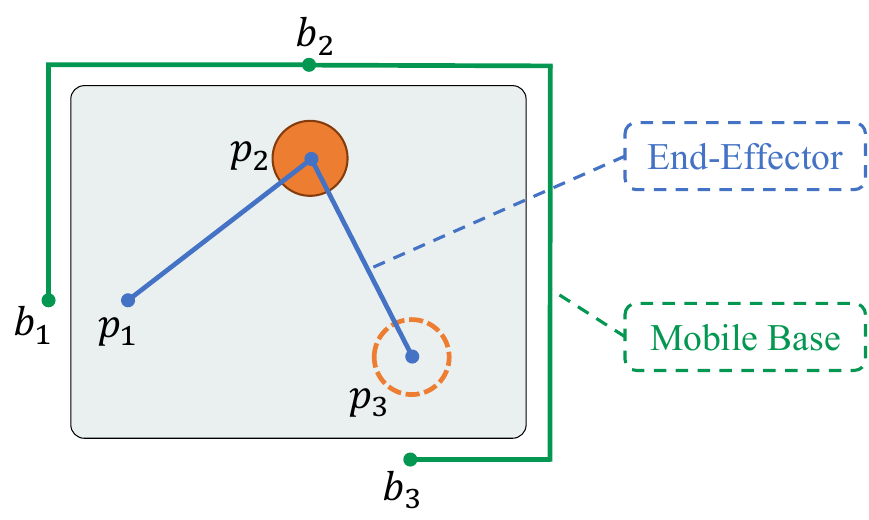}}
	\caption{Visual representation of the travel cost for manipulating the orange object. The blue path indicates end-effector (EE) travel, and the green path indicates mobile base (MB) travel.}
	\label{fig:cost}
\end{figure}

To ensure the physical viability of multi-layer manipulation and transportation, valid stacking configurations within this formulation are governed by three fundamental simplifying assumptions: (i) strict containment, wherein a stacked object is physically bounded by its base (e.g., via container walls) and does not expand the 2D spatial footprint, ensuring that transporting a dynamic stack introduces no additional collision-avoidance dependencies compared to moving the empty base alone; (ii) overhand grasp preservation, dictating that all objects in a stack retain accessible, top-down grasping points; and (iii) a single-item direct stacking capacity, restricting any base to holding a maximum of one immediately stacked item. While this third constraint permits multi-layer transitive stacks (as illustrated in Fig.~\ref{fig:db_example}), limiting direct capacity to one object eliminates complex volumetric calculations and prevents intra-container grasp interference. By enforcing these localized constraints, it is assumed that any permitted multi-layer configuration remains inherently stable and graspable during the transportation of any transitive base.

The objective is to minimize the manipulator's cumulative cost, comprising XY-plane travel distance and a fixed pick-and-place penalty, which reflects real-world operational costs more accurately than discrete action counts. The calculation of this cost depends on the manipulator configuration (Fig.~\ref{fig:sim_env}). For a stationary robot, the cost is based on End-Effector (EE) travel, whereas for a mobile manipulator operating across large workspaces, it is governed by Mobile Base (MB) travel (Fig.~\ref{fig:cost}). Optimizing for this metric is crucial; while a Dynamic Buffer strategy may increase individual action counts, Fig.~\ref{fig:robot} demonstrates that it significantly lowers overall travel cost by transporting grouped objects simultaneously rather than independently.

The cost $c(a^{(t)})$ associated with the $t$-th action, which moves object $o_i$ from $p_i^{(t)}$ to $p_i^{(t+1)}$, is calculated as follows:
\begin{align}
	c(a^{(t)}) &= d(p_m^{(t)}, p_i^{(t)}) + d(p_i^{(t)}, p_i^{(t+1)}) + c_{pp} 
	\label{eq:action_cost}
\end{align}

The planning problem seeks an action sequence $\mathcal{P}$ that minimizes the total cumulative cost $\mathcal{J}(\mathcal{P})$ from initial state $s^{(0)}$ to final state $s^{(T)}$:
\begin{align}
	\min_{\mathcal{P}} \quad & \mathcal{J}(\mathcal{P}) = \sum_{t=0}^{T-1} c(a^{(t)}) + d(p_m^{(T)}, p_m^{(0)}) 
	\label{eq:total_cost} \\
	\text{s.t.} \quad & s^{(t+1)} = \mathcal{T}(s^{(t)}, a^{(t)}), \quad \forall t = 0, \dots, T-1 
	\label{eq:transition}
\end{align}
where:
\begin{itemize}
	\item $a^{(t)}$: Action applied at time $t$;
	\item $d(x, y)$: Manipulator travel distance between position $x$ and position $y$;
	\item $p_m^{(t)}$: Manipulator's position at time $t$;
	\item $p_i^{(t)}$: Object $o_i$'s position at time $t$;
	\item $p_i^{(t+1)}$: Object $o_i$'s next position after $a^{(t)}$;
	\item $c_{pp}$: Constant pick-and-place cost.
\end{itemize}
Critically, the cost in Eq.~\eqref{eq:action_cost} is manipulator-centric: moving a base object $o_i$ incurs a single travel cost, freely relocating all stacked items simultaneously and enabling multi-object manipulation efficiency. The combinatorial optimization problem in Eq.~\eqref{eq:total_cost} is solved using an $A^*$ search (Section~\ref{subsec:search_alg}), which, paired with an admissible heuristic, guarantees optimal solutions within the explored graph.

\section{Scene Rearrangement}\label{sec:method}
This work presents an efficient framework for robotic scene rearrangement that autonomously generates a high-quality action sequence from an initial to a target configuration. The framework comprises scene perception for state representation, search-based planning with a novel heuristic and expansion strategy, and final plan refinement.

\subsection{Scene Perception}
Accurate scene perception is fundamental for robotic manipulation, enabling robots to interpret their environment and the objects within it. This work incorporates 12 object classes (i.e., apple, banana, basket, bowl, box, fork, knife, mug, paper-cup, pear, pot, and spoon) detected using YOLOv5l~\cite{yolov5}. To reconcile raw detections from the initial and target scenes, a \textbf{Consistency Filter} is applied. This filter enforces consistency between views by cleaning up duplicates and matching corresponding objects, generating a single synchronized set of objects that are reliably detected in both scenes. Subsequently, an \textbf{Instance Matching} module establishes final spatial correspondences—critical for classes with multiple instances—by applying the Hungarian algorithm to the Euclidean distances between bounding box centers. This process outputs two aligned and ordered object lists for the planner.

For spatial reasoning, the \textbf{Spatial Relation Inference (SRI)} module builds on the CNN-based architecture from~\cite{ghasemi2024scene}. Trained on a dataset of 604 real tabletop images and 2,500 PyBullet~\cite{coumans2021pybullet} simulated scenes, the CNN processes cropped regions of overlapping object pairs to predict current stacking relations. These relations are assembled into a scene graph representing the initial stacking structure.

\begin{figure}[t]
\centerline{\includegraphics[width=0.48\textwidth]{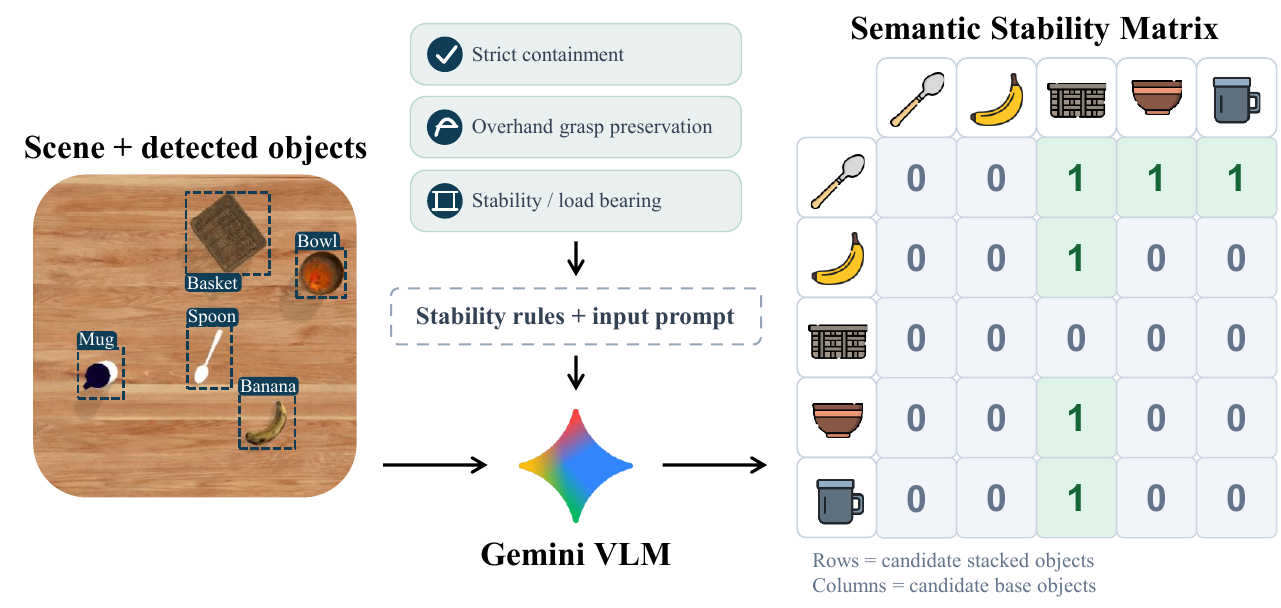}}
\caption{Semantic stability inference pipeline. The full scene image, object crops, and class labels are provided to the Gemini VLM with prompt-based stability constraints. The model outputs a binary stability matrix indicating whether each candidate object (row) can be stably placed in each candidate base (column).}
\label{fig:vlm_stability}
\end{figure}

While the SRI module detects current object relations, the planner requires a comprehensive assessment of multi-layer stacking stability to identify all permissible future configurations. To abstract complex dynamic constraints away from the execution stage, the Gemini \textbf{Vision-Language Model (VLM)}~\cite{gemini} is integrated as an upfront semantic stability evaluator (Fig.~\ref{fig:vlm_stability}). By processing the full scene image, bounding box crops of individual objects, and their corresponding class labels, the VLM evaluates every possible pairwise combination to generate a scenario-specific binary stability matrix, which dictates the valid stacking transitions available to the planner.

Guided by targeted prompts, the VLM acts as a filter to identify proposed pairings that are predicted to satisfy the structural assumptions detailed in Section~\ref{sec:formulation}. Using zero-shot reasoning, the model assesses whether an object can be safely contained within another object while preserving overhand grasp accessibility. This stability evaluation is fundamentally semantic; rather than relying on explicit physical measurements, the VLM leverages broad priors to infer 3D spatial affordances from 2D visual features. By contextualizing an object's geometry with its semantic identity, the model qualitatively assesses relative physical constraints, including containment, mass distribution, and load-bearing capacity. This holistic reasoning enables the model to reject semantically implausible or potentially top-heavy configurations and to evaluate complex spatial interactions. By resolving these transport-related considerations during the perception stage, the resulting stability matrix restricts the planner to stacking pairings predicted to remain stable during transport.

\subsection{Search Algorithm}
\label{subsec:search_alg}
To avoid the computational cost of exhaustive methods, the $A^*$ algorithm is employed to efficiently find high-quality rearrangement plans. This algorithm, building upon its application in~\cite{hu2025strap}, extends its core functionality with action primitives (move and stack) defined in Section~\ref{sec:formulation}. The search space is a directed graph where nodes are states $s$ and edges represent valid transitioning actions. A priority heap accelerates exploration by expanding nodes with the lowest estimated total cost $f(s) = g(s) + h(s)$, where $g(s)$ is the cost-to-come from the initial state, and the heuristic function $h(s)$ estimates the cost-to-go to the target state. Given an admissible heuristic, the search guarantees finding a cost-optimal plan within the explored discrete graph upon expanding the target state $s^{(T)}$; in practical deployment with a finite planning budget, it returns the best solution discovered.

\textbf{Heuristic Function with Stacking Awareness}: A proper heuristic is critical for $A^*$'s efficiency and optimality~\cite{hart1968astar}. The baseline heuristic~\cite{hu2025strap} considers direct goal distances and pick-and-place costs. However, incorporating stacking primitives can make direct moves costlier than traveling as a stack (Fig.~\ref{fig:robot}), causing the baseline to overestimate the true cost. To preserve admissibility, a novel heuristic is introduced that evaluates the minimum cost for each object by comparing direct moves against stacking options. The full heuristic is defined as:
\begin{align}
    \label{eq:heuristic_full}
    h(s^{(t)}) &= \sum_{o_i \in \mathcal{U}^{(t)}} h_i(s^{(t)}) + |\mathcal{U}^{(t)}| \cdot c_{pp}, \\
    \label{eq:heuristic_per_object}
    h_i(s^{(t)}) &= 
    \begin{cases} 
    0, & \text{if } \mathcal{B}^{(t)}_i \cap \mathcal{U}^{(t)} \neq \emptyset \\
    \hat{h}_i(s^{(t)}), & \text{otherwise}
    \end{cases}, \\
    \label{eq:heuristic_unstacked}
    \hat{h}_i(s^{(t)}) &= \min\Big(d(p_i^{(t)}, p_i^{(T)}), \big\{ d(p_i^{(t)}, p_k^{(t)}) + c_{pp} \big\}_{o_k \in \mathcal{O}^{(t)}_i} \Big)
\end{align}
where $\mathcal{U}^{(t)}$ is the set of remaining objects not at their goals. If an object's goal is a satisfied relative stacking relationship, it is excluded from $\mathcal{U}^{(t)}$, even if its base object is still in transit. The candidate bases $\mathcal{O}^{(t)}_i$ include all remaining objects stable for $o_i$ to stack on. The distance $d(x,y)$ represents the EE or MB travel cost (Section~\ref{sec:formulation}). Finally, $\mathcal{B}^{(t)}_i$ denotes the set of objects $o_i$ is directly or transitively stacked on.

To prove admissibility ($h^*(s^{(t)}) \ge h(s^{(t)})$), the optimal cost $h^*(s^{(t)})$ is lower-bounded by ignoring target occupancy and empty-gripper travel, isolating the mandatory independent cost $c^*_i$ to advance each unresolved object $o_i \in \mathcal{U}^{(t)}$:
\begin{equation}
\label{eq:proof_sum}
h^*(s^{(t)}) \ge \sum_{o_i \in \mathcal{U}^{(t)}} c^*_i
\end{equation}
If $o_i$ rests on a movable transitive base ($\mathcal{B}^{(t)}_i \cap \mathcal{U}^{(t)} \neq \emptyset$), the base's movement carries $o_i$ without independent travel cost. However, $o_i$ ultimately requires at least one pick-and-place operation to reach its target. Thus, $c_i^* \ge c_{pp} = h_i(s^{(t)}) + c_{pp}$ (since $h_i(s^{(t)})=0$ here).

Conversely, in the absence of a movable transitive base ($\mathcal{B}^{(t)}_i \cap \mathcal{U}^{(t)} = \emptyset$), $o_i$ must be picked up and either placed at its target or buffered. Direct placement incurs a cost $c_{\text{dir}} = d(p_i^{(t)}, p_i^{(T)}) + c_{pp}$. By the triangle inequality, static buffering is strictly dominated by $c_{\text{dir}}$ due to longer travel and an extra $c_{pp}$ penalty. Alternatively, dynamic buffering on a valid base $o_k \in \mathcal{O}^{(t)}_i$ incurs $c_{\text{dyn}}(k) = d(p_i^{(t)}, p_k^{(t)}) + 2 \cdot c_{pp}$, accounting for the initial placement and a mandatory final move. Minimizing over all valid actions yields:
\begin{align}
    \begin{split}
        c_i^* &\ge \min\Bigg( c_{\text{dir}}, \big\{ c_{\text{dyn}}(k) \big\}_{o_k \in \mathcal{O}^{(t)}_i} \Bigg) \\
        &= \min\Big(d(p_i^{(t)}, p_i^{(T)}), \big\{ d(p_i^{(t)}, p_k^{(t)}) + c_{pp} \big\}_{o_k \in \mathcal{O}^{(t)}_i} \Big) + c_{pp} \\
        &= \hat{h}_i(s^{(t)}) + c_{pp}
    \end{split}
\end{align}

In all cases, $c^*_i \ge h_i(s^{(t)}) + c_{pp}$. Substituting this into Eq.~\ref{eq:proof_sum} establishes admissibility:
\begin{equation}
    h^*(s^{(t)}) \ge \sum_{o_i \in \mathcal{U}^{(t)}} \big( h_i(s^{(t)}) + c_{pp} \big) = h(s^{(t)})
\end{equation}

\textbf{Goal Attempting}:
To improve planning efficiency in long-horizon tasks and ensure plan feasibility at each search step, the planner adopts the \textit{Goal Attempting} mechanism from STRAP~\cite{hu2025strap}. This component serves as a fast solver capable of quickly finding a feasible trajectory from the current state to the goal under a limited time budget. In this paper, Goal Attempting functionality is powered by an algorithm based on the MCTS planner~\cite{labbe2020mcts}, into which the dynamic stacking primitives are integrated. This MCTS component is designed to find a fast feasible plan, and for this purpose, it allows manipulating the same object consecutively. This can lead to plans of lower quality; however, these immediate redundancies are efficiently addressed by the first stage of the refinement process (Section~\ref{subsec:refine}).

\subsection{Expansion Strategy}
The $A^*$ search relies on efficiently generating successor states. For each object, its goal is defined either by a target position or by a stacking relationship with another object in the target scene. The expansion strategy, detailed in Algorithm~\ref{alg:get_valid_actions}, manages the search's branching factor by limiting the total number of successor actions for each object to a predefined \textit{buffer count} $N_{\text{buf}}$. Enforcing this constraint across at most $N$ remaining unplaced objects limits the branching factor to $O(N \cdot N_{\text{buf}})$, mitigating the problem's inherent exponential time complexity. For any unplaced object $o_k$, this action budget is partitioned using a \textit{stacking bias} parameter, $\beta_{\text{stack}} \in (0, 1)$. 

The algorithm first explores stacking opportunities by identifying all valid base objects $O_k$ capable of supporting $o_k$; feasible bases are identified by iterating over all objects in the environment, with stacking stability verified using a precomputed lookup table prior to planning. From these options, it randomly samples up to $N_{\text{stack}}$ targets for \texttt{stack} actions (Lines 5--7), where $N_{\text{stack}} = \max(\lfloor \beta_{\text{stack}} \cdot N_{\text{buf}} \rfloor, 1)$. Subsequently, the remaining budget, $N_{\text{move}}$, is allocated to \texttt{move} actions depending on goal availability. All valid collision-free placements $P_k$ (Line 9) are rapidly validated for a given object's footprint in a single pass using a Summed-Area Table (SAT), or integral image~\cite{viola2001sat}, applied to the multi-label occupancy map. If the object's goal is currently occupied ($p^{(T)}_k \notin P_k$), up to $N_{\text{move}}$ alternative placements are uniformly sampled from $P_k$ (Lines 10--12). Conversely, if the goal is available, the sampling is bypassed to generate a single \texttt{move} action placing $o_k$ directly at its target (Lines 13--15). This dual approach enables the $A^*$ search to evaluate the cost-effectiveness of both direct-to-goal placements and intermediate stacking actions.

\begin{algorithm}[t]
\caption{Expansion Strategy for Generating Successor Actions}
\label{alg:get_valid_actions}

\DontPrintSemicolon
\SetAlgoLined

\KwIn{$S = (s^{(t)}, s^{(T)})$: The current and target arrangements.}
\KwIn{$N_{\text{buf}}$: The number of buffer slots per object.}
\KwResult{$A_{\text{valid}}$: Set of valid actions.}

\BlankLine
$\mathcal{U} \leftarrow \{ o_i \mid o_i \text{ is not at its goal} \}$\;
$A_{\text{valid}} \leftarrow \emptyset$\;
$N_{\text{stack}} \leftarrow \max(\lfloor \beta_{\text{stack}} \cdot N_{\text{buf}} \rfloor, 1)$\;

\For{\textbf{each} $o_k$ \textbf{in} $\mathcal{U}$}{
    $O_k \leftarrow \text{getValidStackBases}(o_k)$\;
    $O'_{k} \leftarrow \text{Sample}(O_k, N_{\text{stack}})$\;
    $A_{k,\text{stack}} \leftarrow \{ \text{stack}(o_k, o_j) \mid o_j \in O'_{k} \}$\;
    \BlankLine
    $N_{\text{move}} \leftarrow N_{\text{buf}} - |A_{k,\text{stack}}|$\;
    $P_k \leftarrow \text{getValidPositions}(o_k)$\;
    
    \If{$p^{(T)}_k \notin P_k$}{
        $P'_{k} \leftarrow \text{Sample}(P_k, N_{\text{move}})$\;
    }
    \Else{
        $P'_{k} \leftarrow \{p^{(T)}_k\}$ 
    }
    
    $A_{k,\text{move}} \leftarrow \{ \text{move}(o_k, p) \mid p \in P'_{k} \}$\;
    \BlankLine
    $A_{\text{valid}} \leftarrow A_{\text{valid}} \cup A_{k,\text{stack}} \cup A_{k,\text{move}}$\;
}
\BlankLine
\Return{$A_{\text{valid}}$}\;
\end{algorithm}

\subsection{Plan Refinement}\label{subsec:refine}
A two-stage refinement process improves initial plan quality by pruning redundant actions and optimizing temporary buffer placements, extending prior methods to accommodate stacking.

\textbf{Redundancy Pruning}: Redundant actions are eliminated while preserving feasibility by removing consecutive operations on the same object (retaining only the final one) and omitting non-consecutive actions that produce no tangible state change.

\begin{figure}[t!]
    \centering
    \begin{subfigure}[t]{0.24\textwidth}
        \centering
        \includegraphics[width=\linewidth]{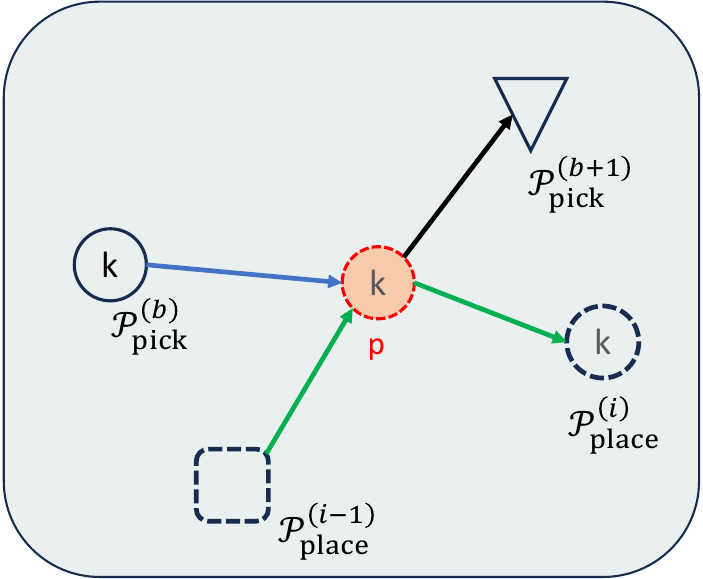}
        \caption{Refinement with a static buffer~\cite{hu2025strap}.}
        \label{fig:ref_static_buffer}
    \end{subfigure}
    \hfill
    \begin{subfigure}[t]{0.24\textwidth}
        \centering
        \includegraphics[width=\linewidth]{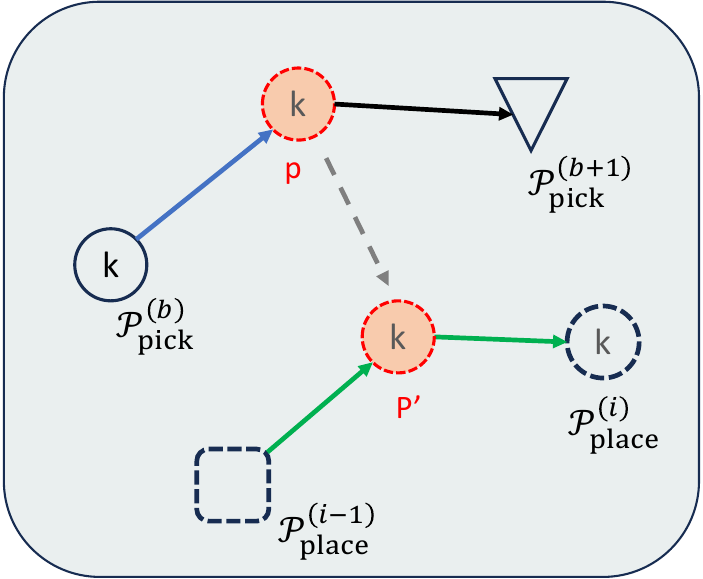}
        \caption{Proposed refinement with a dynamic buffer.}
        \label{fig:ref_dynamic_buffer}
    \end{subfigure}
    \caption{Manipulator trajectories during refinement when moving object $k$ via buffer $p$. Blue paths show travel to $p$ during action $b$; green paths show travel to and from $p$ during action $i$; red marks buffer candidates.}
    \label{fig:refinements}
\end{figure}

\textbf{Buffer Optimality}: The second stage optimizes locations for objects moved multiple times, extending static buffer refinement~\cite{hu2025strap} to include dynamic stacking. For an object $o_k$ buffered at index $b$ and retrieved at index $i$ within an action sequence $\mathcal{P}$, the optimal buffer is selected by minimizing the total manipulator travel distance across four connecting path segments (Fig.~\ref{fig:refinements}).

The optimal position $p^*$ for a \textit{static buffer} minimizes this cost over all valid, collision-free locations $P_k$:
\begin{equation}
	\label{eq:static_buffer_cost}
	\begin{split}
		\min_{p \in P_k} \Big( &d\big(\mathcal{P}^{({b})}_{\text{pick}}, p\big) +
		d\big(p, \mathcal{P}^{({b}+1)}_{\text{pick}}\big) \\
		&+ d\big(\mathcal{P}^{(i-1)}_{\text{place}}, p\big) + d\big(p, \mathcal{P}^{(i)}_{\text{place}}\big) \Big)
	\end{split}
\end{equation}
For a \textit{dynamic buffer}, $o_k$ can be stacked on a base object $o_j$. Accounting for $o_j$'s potential movement between $b$ and $i$, the cost incorporates its position at stacking ($p_j^{(b)}$) and retrieval ($p_j^{(i)}$):
\begin{equation}
	\label{eq:dynamic_buffer_cost}
	\begin{split}
		\min_{o_j \in \mathcal{O}_k} \Big( &d\big(\mathcal{P}^{({b})}_{\text{pick}}, p^{({b})}_j\big) + d\big(p^{({b})}_j, \mathcal{P}^{({b}+1)}_{\text{pick}}\big) \\
		&+ d\big(\mathcal{P}^{(i-1)}_{\text{place}}, p^{(i)}_j\big) + d\big(p^{(i)}_j, \mathcal{P}^{(i)}_{\text{place}}\big) \Big)
	\end{split}
\end{equation}
Here, $\mathcal{O}_k$ denotes the valid base objects for $o_k$ that remain unoccupied between $b$ and $i$. The minimum cost between Eq.~(\ref{eq:static_buffer_cost}) and Eq.~(\ref{eq:dynamic_buffer_cost}) dictates the update to action $b$, assigning $o_k$ to either the optimal static buffer $p^*$ or the optimal dynamic buffer $o_j^*$.

\section{Results and Discussion}\label{sec:results}
This section presents the experimental validation and analysis of the proposed stacking approach for object rearrangement planning. The performance of the proposed system is evaluated in four parts: a comprehensive algorithm comparison against several baselines, an analysis of key components and parameters within the framework, an ablation study, and a final full pipeline experiment conducted in both simulation and the real world.

\subsection{Algorithm Comparison}
To validate the effectiveness of the Dynamic-Stacking (DS) approach, its performance is compared against several baseline methods, representing both traditional Non-Stacking (NS) and Static-Stacking (SS) strategies. The following algorithms are evaluated:
\\\textbf{MCTS}: The standard MCTS planner~\cite{labbe2020mcts}, used for finding feasible rearrangement plans.
\\\textbf{MCTS+DS}: The proposed DS approach integrated into the MCTS planner.
\\\textbf{STRAP}: The \(A^*\) planner from~\cite{hu2025strap}, used for planning high-quality rearrangement sequences. This planner employs a Goal Attempting mechanism, for which an MCTS planner is utilized internally.
\\\textbf{STRAP+SS}: The SS approach, representing existing stacking methods (e.g., as conceptualized in ORLA*~\cite{gao2025orla}), integrated into the STRAP planner.
\\\textbf{STRAP+DS}: The proposed DS approach integrated into the STRAP planner.

For a fair comparison with baselines, which do not support pre-existing stacking, initial and target arrangements are generated without stacking relations. In this context, stacking primitives are exclusively used for intermediate buffering during the planning process. For each object count $n$, 40 random initial-target scene pairs are generated, and one trial is executed for each pair. Each trial is subject to a 6-minute time limit to find a cost-efficient solution. Regardless of the success rate, the reported performance metrics—average travel cost, average number of actions, and average computation time—are evaluated over successful trials to remain consistent with previous work~\cite{hu2025strap}.
Travel cost is considered the primary metric, as it comprehensively quantifies both manipulator travel distances and pick-and-place operations.
To ensure a consistent experimental setup, uniformly sized objects are used. The scene density $\varphi$ is then defined as the ratio of the total object footprint area $\sum_{o \in \mathcal{O}} S(o)$ to the total table area $S_T$~\cite{gao2025orla}, where $S(o)$ is the footprint size of object $o$. The table dimensions are $1 \times 1$ meters for stationary mode and $1 \times 2$ meters for mobile mode, and manipulator movement costs are normalized by the width of the table. Furthermore, a pick-and-place cost \( c_{pp} \) of 0.2 (one-fifth of the cost of traversing the table width) is applied per action.

\begin{figure}[t]
	\centerline{\includegraphics[ width = 0.5\textwidth]{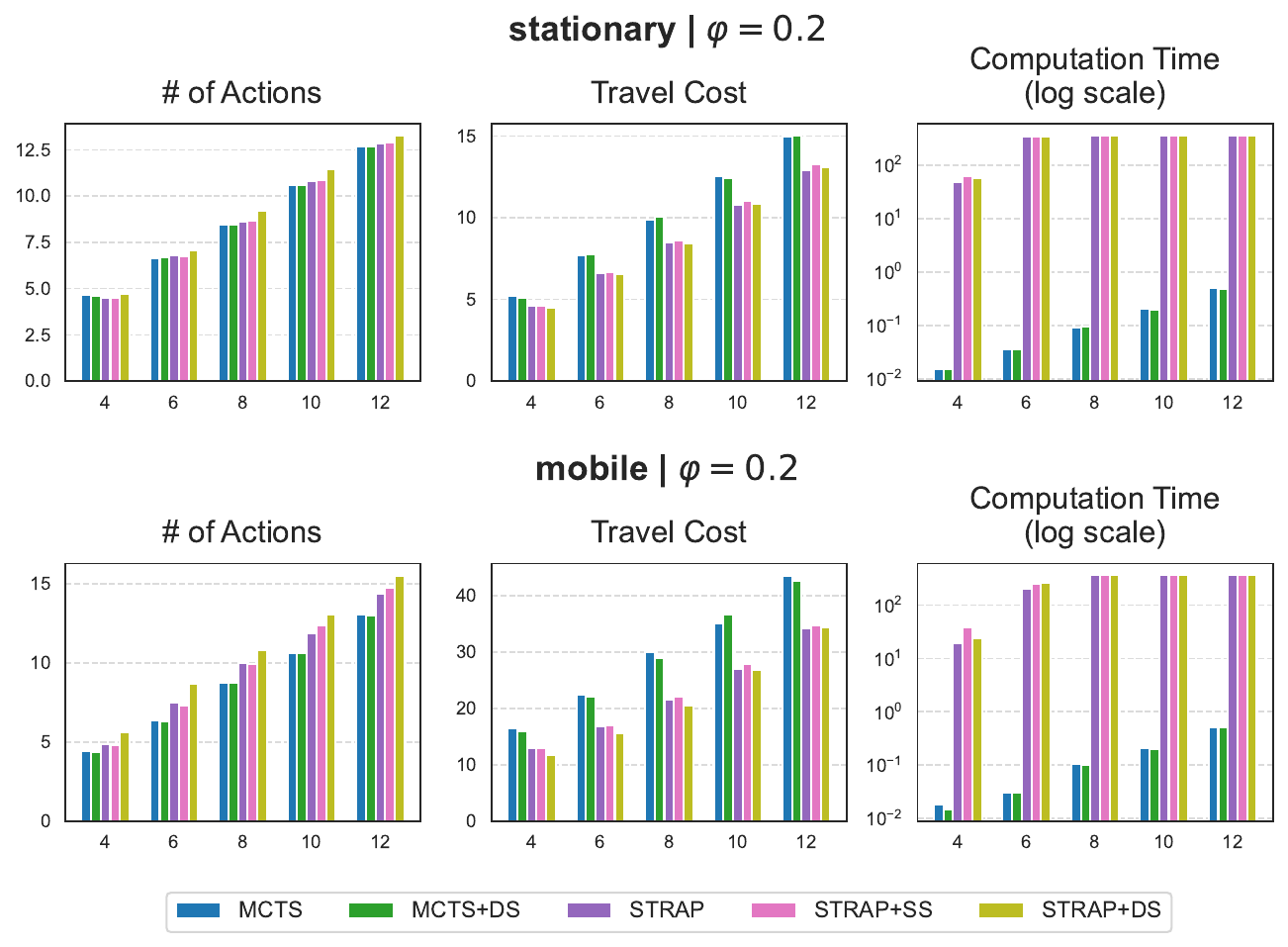}}
	\caption{Results for low-density scenes ($\varphi=0.2$). The horizontal axis represents the number of objects in the scene. The top row illustrates performance for the stationary manipulator, while the bottom row presents the corresponding metrics for the mobile manipulator.}
	\label{fig:results_phi0.2}
\end{figure}

\begin{figure}[t]
	\centerline{\includegraphics[ width = 0.5\textwidth]{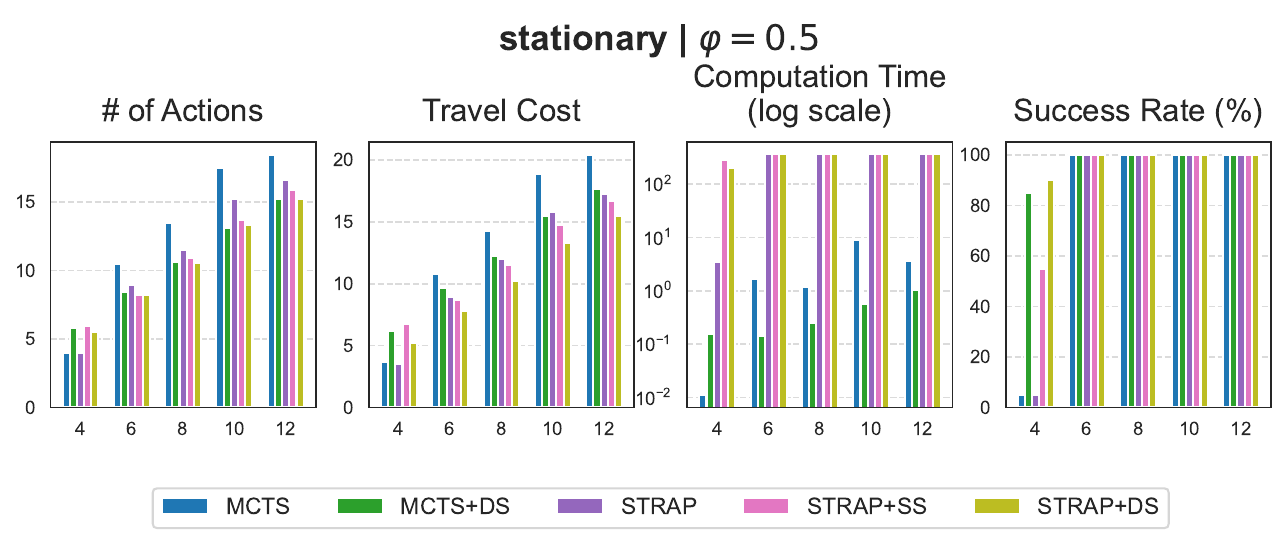}}
	\caption{Results for high-density scenes ($\varphi=0.5$). The horizontal axis represents the number of objects in the scene. The plots display algorithm performance for the stationary manipulator in these challenging environments.}
	\label{fig:results_phi0.5}
\end{figure}

To provide a robust assessment of performance that accounts for varying success rates and to enable a direct comparison of algorithms across different numbers of objects, the Expected Successful Cost (ESC) is defined as the average cost of successful trials divided by the success rate:
\begin{equation}
    \text{ESC}(n) = \frac{\text{Average Cost of Successful Trials}(n)}{\text{Success Rate}(n)}
    \label{eq:esc}
\end{equation}
where the success rate is represented as a fraction in $(0,1]$. For an aggregate measure across all evaluated object counts, the Overall Performance Score (OPS) of an algorithm is defined as the geometric mean of its ESC values over all relevant $n$:
\begin{equation}
	\text{OPS} = \left( \prod_{n \in \mathcal{N}} \text{ESC}(n) \right)^{\frac{1}{|\mathcal{N}|}}
	\label{eq:ops}
\end{equation}
where $\mathcal{N}$ is the set of evaluated object counts. To quantify the performance improvement of algorithm B over algorithm A, the Performance Improvement Rate (PIR) based on the OPS metric is calculated as:
\begin{equation}
	\text{PIR}(B \text{ over } A) = \left( 1 - \frac{\text{OPS}(B)}{\text{OPS}(A)} \right) \times 100\%
	\label{eq:pir}
\end{equation}

\begin{table}[tp]
	\centering
	\caption{Comparison of algorithm performance across different scenarios. OPS and PIR are presented for each algorithm, covering stationary (EE) and mobile (MB) manipulators in low ($\varphi=0.2$) and high ($\varphi=0.5$) density scenes.}
	\label{tab:results}
	\resizebox{\columnwidth}{!}{%
		\begin{tabular}{@{}ccc|cc|ccc@{}}
			& & & \multicolumn{2}{c|}{MCTS} & \multicolumn{3}{c}{STRAP} \\
			\hhline{~~~|--|---} 
			Manip. & $\varphi$ & Metrics & NS     & DS      & NS     & SS       & DS       \\ 
			\hhline{-|-|-|-|-|-|-|-}
			\multirow{2}{*}{EE}  & \multirow{2}{*}{0.2}
			& OPS   & 9.43   & 9.42    & 8.15   & 8.27     & 8.12     \\
			&& PIR   & - & 0.15\%  & - & -1.40\%  & 0.41\%   \\
			\hhline{-|-|-|-|-|-|-|-}
			\multirow{2}{*}{MB} & \multirow{2}{*}{0.2}
			& OPS   & 27.94  & 27.56   & 21.25  & 21.59     & 20.33    \\
			&& PIR   & - & 1.35\%  & - & -1.59\%  & \textbf{4.34}\%   \\
			\hhline{-|-|-|-|-|-|-|-}
			\multirow{2}{*}{EE} & \multirow{2}{*}{0.5}
			& OPS   & 15.62  & 13.39   & 13.07   & 12.52     & 11.31     \\
			&& PIR   & - & 14.24\% & - & 4.20\%   & \textbf{13.48}\%  \\ 
			\bottomrule
		\end{tabular}%
	}
\end{table}

The performance comparison across all algorithms and scenarios is shown in a combination of figures and a table. The results in Fig.~\ref{fig:results_phi0.2} and Fig.~\ref{fig:results_phi0.5} illustrate the performance trends, while Table~\ref{tab:results} provides a numerical summary of the OPS and PIR for each algorithm.
Fig.~\ref{fig:results_phi0.2} shows results of low-density scenes for both stationary and mobile manipulators, while the computationally expensive nature of high-density scenes limits the analysis in Fig.~\ref{fig:results_phi0.5} to the stationary manipulator only. Notably, all algorithms achieved a 100\% success rate in low-density scenarios.
For the STRAP-based methods, this performance is aided by the Goal Attempting mechanism; consequently, ESC equals the average trial cost in these cases.

For low-density scenarios (\(\varphi=0.2\)) in Fig.~\ref{fig:results_phi0.2}, adding dynamic stacking to MCTS (MCTS+DS) does not yield a significant improvement in performance compared to baseline MCTS, as MCTS primarily focuses on feasibility, and complex buffering is less critical in low-density settings. For the stationary manipulator (Fig.~\ref{fig:results_phi0.2} top), STRAP+DS achieves a 0.41\% cost improvement over baseline STRAP. This is reflected in the OPS values in Table~\ref{tab:results} (8.15 for STRAP vs. 8.12 for STRAP+DS). In contrast, for the mobile manipulator (Fig.~\ref{fig:results_phi0.2} bottom), the benefits of dynamic stacking are more pronounced. STRAP+DS achieves a 4.34\% cost improvement over baseline STRAP, as seen in Table~\ref{tab:results} (21.25 for STRAP vs. 20.33 for STRAP+DS). 
This improvement in cost for STRAP+DS in both stationary and mobile scenarios is accompanied by an increase in the average number of actions. This trend is expected, as in larger mobile workspaces, grouping objects via dynamic stacking and transferring them together can be more cost-efficient despite requiring more individual actions. In both stationary and mobile scenarios, STRAP+SS slightly degrades performance compared to baseline STRAP.

High-density scenarios (\(\varphi=0.5\)) pose significant challenges, particularly for algorithms lacking stacking capabilities, as free buffer space can become extremely scarce, rendering some problems infeasible. It should be noted that while the work of~\cite{hu2025strap} did not evaluate performance in such high-density environments, this paper's algorithms are specifically tested under these conditions.
In these challenging environments, the limited available space severely impacts success rates for non-stacking algorithms. For instance, the low average cost for STRAP at \( n=4 \) is misleading, as it is derived from only 5\% successful trials (2 out of 40 scenarios). Therefore, OPS is calculated across $n \in [5,12]$, excluding the 4-object scenario from aggregate metrics because the near-zero success rate of non-stacking methods disproportionately distorts comparative evaluation.

From Fig.~\ref{fig:results_phi0.5}, it can be inferred that stacking methods generally outperform non-stacking methods in high-density scenarios, which is supported by the consistently lower OPS values in Table~\ref{tab:results}. While STRAP+SS provides a PIR of 4.20\% over the STRAP baseline, STRAP+DS demonstrates even greater advantages by achieving both lower travel costs and fewer actions. Most notably, STRAP+DS yields a substantial PIR of 13.48\% over the baseline, with its final OPS (11.31) being the lowest among all planners. This indicates that dynamic stacking is substantially beneficial in high-density rearrangement tasks.

\subsection{Component and Parameter Analysis}
An analysis of the core components and hyperparameters is conducted to quantify their contributions and detail the selection of their configurations. The analysis focuses first on parameter sensitivity and subsequently on the impact of post-processing refinement.

\subsubsection{Parameter Sensitivity Analysis}
\begin{figure}[t]
	\centerline{\includegraphics[width=0.5\textwidth]{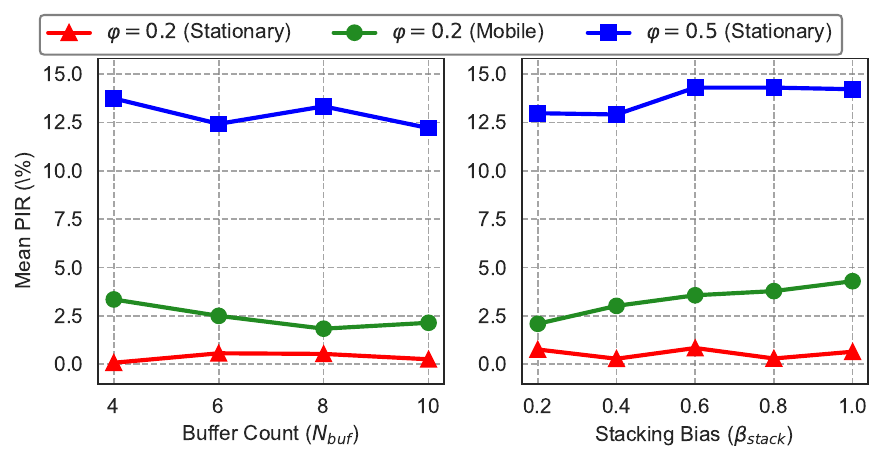}}
	\caption{Parameter sensitivity analysis evaluating the mean PIR of STRAP+DS over STRAP. Left: Effect of varying buffer count ($N_{\text{buf}}$). Right: Impact of stacking bias ($\beta_{\text{stack}}$) given the optimally tuned $N_{\text{buf}}$ for each respective configuration.}
	\label{fig:param_sensitivity}
\end{figure}

A sensitivity study evaluates the impact of buffer count ($N_{\text{buf}}$) and stacking bias ($\beta_{\text{stack}}$) on the mean PIR of STRAP+DS over STRAP. Optimal parameters were tuned sequentially using a one-quarter validation split: first maximizing PIR over $N_{\text{buf}}$, then tuning $\beta_{\text{stack}}$. As Fig.~\ref{fig:param_sensitivity} shows, optimal configurations are environment- and manipulator-dependent. Peak settings ($N_{\text{buf}}, \beta_{\text{stack}}$) are (6, 0.6) for stationary manipulators at low density, (4, 1.0) for mobile manipulators at low density, and (4, 0.6) for stationary manipulators at high density. This highlights the need to carefully balance search scope and plan quality: while increasing $N_{\text{buf}}$ expands the search space, it can degrade plan quality given a fixed search time limit. Conversely, adjusting $\beta_{\text{stack}}$ merely alters the proportional allocation of stack buffers without directly affecting the total search space size. These tuned configurations were applied in the preceding main evaluations.

\subsubsection{Refinement Study}
Table~\ref{tab:refinement} details the impact of plan refinement, reporting the refined OPS in parentheses and the PIR over the unrefined baseline. 
In low-density scenarios ($\varphi=0.2$), dynamic refinement yields marginal gains over static refinement, suggesting that stacking provides negligible advantage when ample free space is available. However, in high-density scenarios ($\varphi=0.5$), dynamic refinement consistently provides higher PIRs. For instance, applying dynamic refinement to MCTS achieves a 12.58\% improvement, substantially outperforming the 7.30\% from static refinement.
Notably, while STRAP+DS exhibits lower PIR values, its final OPS remains superior. This indicates that its primary search already incorporates efficient stacking strategies, yielding high-quality initial plans with less room for post-processing optimization.

\begin{table}[tp]
 \centering
 \caption{Comparison of travel-cost improvements achieved by applying refinement methods. Values are presented as (OPS) and PIR\% for each algorithm, where OPS is the overall performance score after refinement. ``Static'' rows show PIR from applying refinement with only static buffers (based on~\cite{hu2025strap}); ``Dynamic'' rows show PIR from applying full refinement (incorporating dynamic buffers).
}
 \label{tab:refinement}
 \resizebox{\columnwidth}{!}{%
  \begin{tabular}{@{}ccc|ccccc@{}}
   & & & \multicolumn{2}{c}{MCTS} & \multicolumn{3}{c}{STRAP} \\
   \hhline{~~~--||---} 
   Manip. & $\varphi$ & Refine & NS & DS  & NS & SS  & DS  \\ 
   \hhline{-|-|-|-|-|-|-|-}
   \multirow{3}{*}{EE} & \multirow{3}{*}{0.2}
   & Static & \begin{tabular}[c]{@{}c@{}}(8.94)\\5.28\%\end{tabular} & \begin{tabular}[c]{@{}c@{}}(8.96)\\4.91\%\end{tabular} & \begin{tabular}[c]{@{}c@{}}(8.07)\\1.07\%\end{tabular} & \begin{tabular}[c]{@{}c@{}}(8.17)\\1.17\%\end{tabular}  & \begin{tabular}[c]{@{}c@{}}(8.07)\\0.67\%\end{tabular}  \\
   \hhline{~~-|-|-|-|-|-}
   & & Dynamic & \begin{tabular}[c]{@{}c@{}}(8.92)\\5.44\%\end{tabular} & \begin{tabular}[c]{@{}c@{}}(8.95)\\4.96\%\end{tabular} & \begin{tabular}[c]{@{}c@{}}(8.06)\\1.18\%\end{tabular} & \begin{tabular}[c]{@{}c@{}}(8.17)\\1.23\%\end{tabular}  & \begin{tabular}[c]{@{}c@{}}(8.06)\\0.68\%\end{tabular}  \\
   \hhline{-|-|-|-|-|-|-|-}

   \multirow{3}{*}{MB} & \multirow{3}{*}{0.2}
   & Static & \begin{tabular}[c]{@{}c@{}}(26.24)\\6.08\%\end{tabular} & \begin{tabular}[c]{@{}c@{}}(26.18)\\5.02\%\end{tabular} & \begin{tabular}[c]{@{}c@{}}(20.73)\\2.46\%\end{tabular} & \begin{tabular}[c]{@{}c@{}}(20.95)\\2.95\%\end{tabular}  & \begin{tabular}[c]{@{}c@{}}(19.94)\\1.88\%\end{tabular}  \\
   \hhline{~~-|-|-|-|-|-}
   & & Dynamic & \begin{tabular}[c]{@{}c@{}}(26.19)\\6.25\%\end{tabular} & \begin{tabular}[c]{@{}c@{}}(26.14)\\5.16\%\end{tabular} & \begin{tabular}[c]{@{}c@{}}(20.59)\\3.08\%\end{tabular} & \begin{tabular}[c]{@{}c@{}}(20.82)\\3.55\%\end{tabular}  & \begin{tabular}[c]{@{}c@{}}(19.92)\\1.99\%\end{tabular}  \\
   \hhline{-|-|-|-|-|-|-|-}

   \multirow{3}{*}{EE} & \multirow{3}{*}{0.5}
   & Static & \begin{tabular}[c]{@{}c@{}}(14.48)\\7.30\%\end{tabular} & \begin{tabular}[c]{@{}c@{}}(12.83)\\4.22\%\end{tabular} & \begin{tabular}[c]{@{}c@{}}(12.67)\\3.09\%\end{tabular} & \begin{tabular}[c]{@{}c@{}}(12.25)\\2.19\%\end{tabular}  & \begin{tabular}[c]{@{}c@{}}(11.21)\\0.94\%\end{tabular}  \\
   \hhline{~~-|-|-|-|-|-}
   & & Dynamic & \begin{tabular}[c]{@{}c@{}}(13.65)\\12.58\%\end{tabular} & \begin{tabular}[c]{@{}c@{}}(12.48)\\6.82\%\end{tabular} & \begin{tabular}[c]{@{}c@{}}(12.30)\\5.94\%\end{tabular} & \begin{tabular}[c]{@{}c@{}}(11.92)\\4.86\%\end{tabular}  & \begin{tabular}[c]{@{}c@{}}(11.17)\\1.22\%\end{tabular}  \\
   \bottomrule
  \end{tabular}%
 }
\end{table}

\subsection{Ablation Study}
This study evaluates the individual contributions of two core system components: the impact of synthetic training data on the SRI module's robustness, and the efficacy of post-processing refinement stages in reducing trajectory costs.

\subsubsection{SRI Robustness}
To establish the robustness boundaries of the SRI module, its base perception accuracy is first quantified, followed by an evaluation of the impact of synthetic training data. The baseline module achieved an accuracy of 98.26\% on the simulation test set and 92.42\% on the real test set, demonstrating reliable real-world spatial-relation inference under the evaluated conditions. However, when trained exclusively on real-world data without synthetic data augmentation, the real-world accuracy dropped to 87.77\%. This degradation highlights the critical role of synthetic data in bridging the reality gap and providing the necessary generalization for robust real-world deployment.

\subsubsection{Refinement Stages}
Table~\ref{tab:refinement_ablation} quantifies the contributions of post-planning Redundancy Pruning (RP) and Buffer Optimization (BO) to total cost reduction. This evaluation strictly targets the final post-processing RP, as internal RP within the A* search remains active. Results show BO is the primary driver of cost reduction. Post-processing RP yields minimal standalone improvement because the A* expansion naturally prevents consecutive actions on a single object. However, combining both stages maximizes efficiency. In high-density scenarios ($\varphi=0.5$), the combined reduction (4.92\%) exceeds the sum of individual contributions (4.87\%), demonstrating a synergistic effect: removing residual redundant actions via RP unlocks deeper optimization opportunities for BO.

\begin{table}[tp]
 \centering
 \caption{Ablation on plan refinement: individual and combined effects of Redundancy Pruning (RP) and Buffer Optimization (BO) on travel cost reduction.}
 \label{tab:refinement_ablation}
 \resizebox{0.7\columnwidth}{!}{%
  \begin{tabular}{@{}cc|c|c|c@{}}
   Manip. & $\varphi$ & RP & BO & RP+BO \\ 
   \hhline{-|-|-|-|-}
   EE & 0.2 & 0.02\% & 2.63\% & 2.66\% \\
   MB & 0.2 & 0.02\% & 3.82\% & 3.84\% \\
   EE & 0.5 & 0.35\% & 4.52\% & 4.92\% \\
   \bottomrule
  \end{tabular}
 }
\end{table}

\subsection{Full Pipeline Experiment}
The full pipeline, from visual input to robot execution, was evaluated in a more realistic setting than the algorithm comparisons. While the algorithm comparisons utilized uniformly sized objects to isolate planning complexity, these full-pipeline experiments validate the system's robustness by utilizing a diverse set of non-uniform objects with varying sizes, alongside scenarios containing pre-existing stacking relationships. The system's inputs were limited to two RGB images representing the initial and target scenes. 

\subsubsection{Simulation Results}
The system was evaluated on 40 rearrangement tasks in a simulation environment, with each task containing 10 randomly selected objects. These experiments were conducted using a Franka Emika Panda manipulator model~\cite{gaz2019panda} within the PyBullet physics engine~\cite{coumans2021pybullet}. The system successfully solved 39 out of 40 tasks, utilizing an average of 7.6 actions for successful runs.

\begin{figure}[t]
	\centerline{\includegraphics[width=0.5\textwidth]{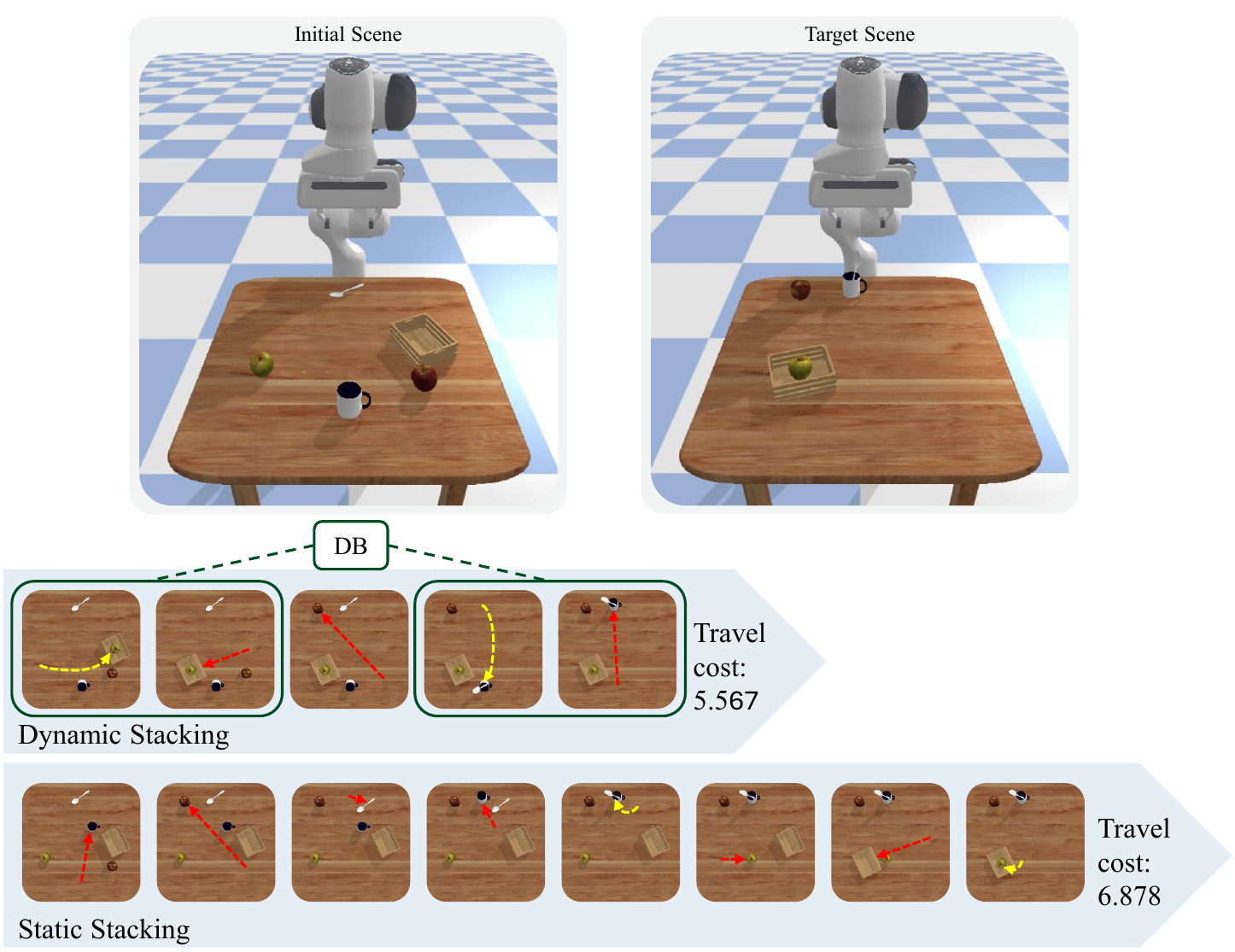}}
	\caption{A comparison of rearrangement plans using Static vs. Dynamic Stacking for a scene with pre-existing stacks. Dynamic Stacking allows objects like the box and mug to be used as movable ``dynamic buffers'', resulting in a more efficient, lower-cost plan as shown.}
	\label{fig:ds_ss}
\end{figure}

A specific simulation was examined to illustrate the fundamental difference between the dynamic and static stacking approaches, particularly in scenarios with pre-existing stacks. Static stacking primitives do not permit the relocation of an object that serves as a base for another, which constrains the available planning options. In contrast, the dynamic stacking approach allows an entire stack to be moved by manipulating its base object. This key capability is highlighted in Fig.~\ref{fig:ds_ss}. The figure demonstrates how leveraging an existing stack as a single movable unit can lead to a more efficient plan with a significantly lower travel cost compared to the sequence of actions available to a static stacking approach.

\subsubsection{Real-World Results}
\begin{figure}[t]
	\centerline{\includegraphics[width=0.4\textwidth]{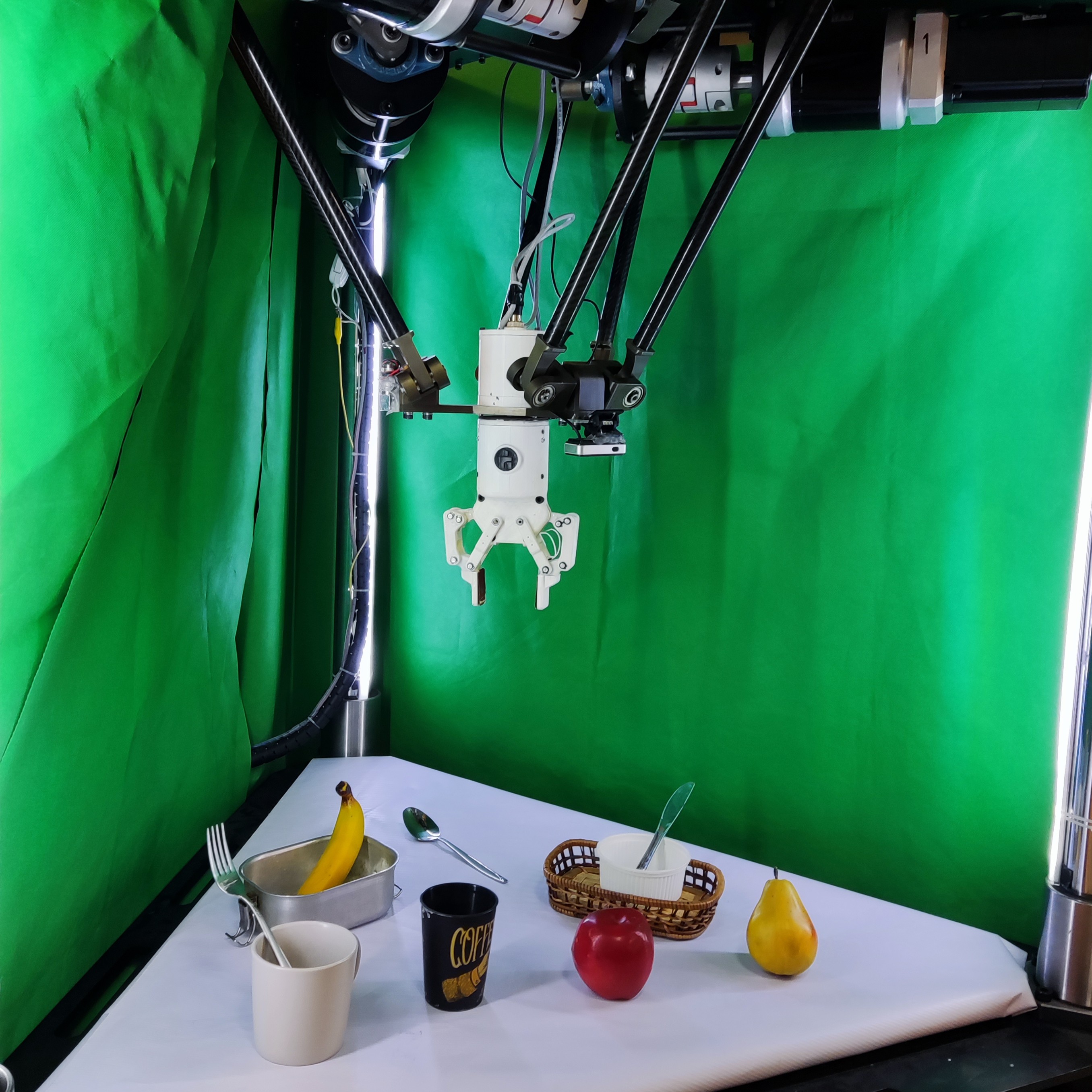}}
	\caption{The real-world experimental setup. The Delta robot is shown preparing to perform a rearrangement task on the tabletop.}
	\label{fig:real_robot}
\end{figure}

Real-robot experiments were performed using a custom-built Delta robot described in~\cite{dastjerdi2020delta}. An overview of the real-world setup is shown in Fig.~\ref{fig:real_robot}. The system was evaluated on 20 rearrangement tasks, each composed of 5 objects. Of the 20 experiments, 18 were successful. For successful tasks, the system used an average of 5.1 actions for execution. 
The few failures observed in both the simulation and real-world experiments stemmed from a common source, namely errors in the visual perception system. These issues typically involved incorrectly estimated object counts, mismatches between source and target objects, or the misclassification of stacking relations.

Code and supplementary videos are available on GitHub.\footnote{\url{https://github.com/armanbarghi/DBRP-Code}}

\section{Conclusion}\label{sec:conclusion}
This paper presented a novel approach to tabletop rearrangement planning by introducing the Dynamic Buffer concept. Inspired by human grouping strategies, this method enables multi-object manipulation by allowing temporary object stacks to be dynamically relocated, thereby providing flexible, non-fixed buffer locations. Incorporating dynamic stacking into existing rearrangement planners improves plan quality and reduces manipulator travel cost. Experimental evaluations demonstrated cost reductions of 13.48\% in challenging, high-density stationary-manipulator scenarios and 4.34\% for low-density rearrangement over large workspaces using a mobile base. These findings highlight dynamic buffering as an effective strategy for improving feasibility in space-constrained environments and reducing planning cost even when stacking is not strictly required. Finally, integrating learning-based heuristics into the $A^*$ search framework represents a promising future direction for accelerating search convergence and scaling to more complex, cluttered scenes.

\bibliographystyle{IEEEtran}
\bibliography{bibliography}

@IEEEtranBSTCTL{IEEEexample:BSTcontrol,
  CTLuse_forced_etal       = "yes",
  CTLmax_names_forced_etal = "6",
  CTLnames_show_etal       = "1"
}

@article{wu2023tidybot,
  author = {Wu, Jimmy and Antonova, Rika and Kan, Adam and Lepert, Marion and Zeng, Andy and Song, Shuran and Bohg, Jeannette and Rusinkiewicz, Szymon and Funkhouser, Thomas},
  title = {{TidyBot}: Personalized Robot Assistance with Large Language Models},
  journal = {Auton. Robots},
  volume = {47},
  number = {8},
  pages={1087--1102},
  year = {2023},
}

@article{garrett2021tamp,
  author = {Garrett, Caelan Reed and Chitnis, Rohan and Holladay, Rachel and Kim, Beomjoon and Silver, Tom and Kaelbling, Leslie Pack and Lozano-P{\'e}rez, Tom{\'a}s},
  title = {Integrated Task and Motion Planning},
  journal = {Annu. Rev. Control Robot. Auton. Syst.},
  volume = {4},
  number = {1},
  pages = {265--293},
  year = {2021},
}

@article{hart1968astar,
  author={Hart, Peter E. and Nilsson, Nils J. and Raphael, Bertram},
  journal={IEEE Trans. Syst. Sci. Cybern.}, 
  title={A Formal Basis for the Heuristic Determination of Minimum Cost Paths}, 
  volume={4},
  number={2},
  pages={100--107},
  year={1968},
}

@inproceedings{srinivas2023multiobj,
  author={Srinivas, Kishore and Ganti, Shreya and Parikh, Rishi and Ahmad, Ayah and Agboh, Wisdom and Dogar, Mehmet and Goldberg, Ken},
  booktitle={Proc. IEEE 19th Int. Conf. Autom. Sci. Eng. (CASE)}, 
  title={The {Busboy} Problem: Efficient Tableware Decluttering Using Consolidation and Multi-Object Grasps}, 
  year={2023},
  pages={1--6},
}

@inproceedings{wu2023multiobj,
  author={Wu, Zewen and Tang, Jian and Chen, Xingyu and Ma, Chengzhong and Lan, Xuguang and Zheng, Nanning},
  booktitle={Proc. IEEE/RSJ Int. Conf. Intell. Robots Syst. (IROS)}, 
  title={Prioritized Planning for Target-Oriented Manipulation via Hierarchical Stacking Relationship Prediction}, 
  year={2023},
  pages={4873--4880},
}

@inproceedings{agboh2023multiobj,
  author={Agboh, Wisdom C. and Sharma, Satvik and Srinivas, Kishore and Parulekar, Mallika and Datta, Gaurav and Qiu, Tianshuang and Ichnowski, Jeffrey and Solowjow, Eugen and Dogar, Mehmet and Goldberg, Ken},
  booktitle={Proc. IEEE/RSJ Int. Conf. Intell. Robots Syst. (IROS)}, 
  title={Learning to Efficiently Plan Robust Frictional Multi-Object Grasps}, 
  year={2023},
  pages={10660--10667},
}

@article{han2018toro,
  author = {Han, Shuai D and Stiffler, Nicholas M and Krontiris, Athanasios and Bekris, Kostas E and Yu, Jingjin},
  title = {Complexity Results and Fast Methods for Optimal Tabletop Rearrangement with Overhand Grasps},
  journal = {Int. J. Robot. Res.},
  volume = {37},
  number = {13-14},
  pages = {1775--1795},
  year = {2018},
}

@article{labbe2020mcts,
  author={Labb{\'e}, Yann and Zagoruyko, Sergey and Kalevatykh, Igor and Laptev, Ivan and Carpentier, Justin and Aubry, Mathieu and Sivic, Josef},
  title={{Monte-Carlo} Tree Search for Efficient Visually Guided Rearrangement Planning}, 
  journal={IEEE Robot. Autom. Lett.}, 
  volume={5},
  number={2},
  pages={3715--3722},
  year={2020},
}

@inproceedings{zhu2021symbolic,
  author={Zhu, Yifeng and Tremblay, Jonathan and Birchfield, Stan and Zhu, Yuke},
  booktitle={Proc. IEEE Int. Conf. Robot. Autom. (ICRA)},
  title={Hierarchical planning for long-horizon manipulation with geometric and symbolic scene graphs},
  year={2021},
  pages={6541--6548},
}

@inproceedings{huang2024pmmr,
  author={Huang, Baichuan and Zhang, Xujia and Yu, Jingjin},
  booktitle={Proc. IEEE Int. Conf. Robot. Autom. (ICRA)}, 
  title={Toward Optimal Tabletop Rearrangement with Multiple Manipulation Primitives}, 
  year={2024},
  pages={10860--10866},
}

@inproceedings{song2020sort,
  author={Song, Haoran and Haustein, Joshua A. and Yuan, Weihao and Hang, Kaiyu and Wang, Michael Yu and Kragic, Danica and Stork, Johannes A.},
  booktitle={Proc. IEEE/RSJ Int. Conf. Intell. Robots Syst. (IROS)}, 
  title={Multi-Object Rearrangement with {Monte Carlo} Tree Search: A Case Study on Planar Nonprehensile Sorting}, 
  year={2020},
  pages={9433--9440},
}

@article{gao2023runningbuf,
  author = {Kai Gao and Si Wei Feng and Baichuan Huang and Jingjin Yu},
  title ={Minimizing running buffers for tabletop object rearrangement: Complexity, fast algorithms, and applications},
  journal = {Int. J. Robot. Res.},
  volume = {42},
  number = {10},
  pages = {755--776},
  year = {2023},
}

@inproceedings{gao2022trlb,
  author={Gao, Kai and Lau, Darren and Huang, Baichuan and Bekris, Kostas E. and Yu, Jingjin},
  booktitle={Proc. IEEE Int. Conf. Robot. Autom. (ICRA)}, 
  title={Fast High-Quality Tabletop Rearrangement in Bounded Workspace}, 
  year={2022},
  pages={1961--1967},
}

@inproceedings{chang2024lgmcts,
  author={Chang, Haonan and Gao, Kai and Boyalakuntla, Kowndinya and Lee, Alex and Huang, Baichuan and Yu, Jingjin and Boularias, Abdeslam},
  booktitle={Proc. IEEE/RSJ Int. Conf. Intell. Robots Syst. (IROS)}, 
  title={{LGMCTS}: Language-Guided {Monte-Carlo} Tree Search for Executable Semantic Object Rearrangement}, 
  year={2024},
  pages={13607--13612},
}

@inproceedings{lou2023hetgnn,
  author={Lou, Xibai and Yu, Houjian and Worobel, Ross and Yang, Yang and Choi, Changhyun},
  booktitle={Proc. IEEE/RSJ Int. Conf. Intell. Robots Syst. (IROS)}, 
  title={Adversarial Object Rearrangement in Constrained Environments with Heterogeneous Graph Neural Networks}, 
  year={2023},
  pages={1008--1015},
}

@article{xu2024see,
  author={Xu, Kechun and Zhou, Zhongxiang and Wu, Jun and Lu, Haojian and Xiong, Rong and Wang, Yue},
  title={Grasp, See, and Place: Efficient Unknown Object Rearrangement With Policy Structure Prior}, 
  journal={IEEE Trans. Robot.}, 
  volume={41},
  pages={464--483},
  year={2025},
}

@inproceedings{qureshi2021nerp, 
  author = {Ahmed H Qureshi and Arsalan Mousavian and Chris Paxton and Michael Yip and Dieter Fox}, 
  booktitle = {Proc. Robotics: Sci. Syst.}, 
  title = {{NeRP}: Neural Rearrangement Planning for Unknown Objects}, 
  year = {2021}, 
  address = {Virtual},
  month = {July}, 
}

@article{kee2025tsmcts,
  author={Kee, Hogun and Oh, Wooseok and Kang, Minjae and Ahn, Hyemin and Oh, Songhwai},
  title={Tidiness Score-Guided {Monte Carlo} Tree Search for Visual Tabletop Rearrangement}, 
  journal={IEEE Robot. Autom. Lett.}, 
  volume={10},
  number={10},
  pages={10090--10097},
  year={2025},
}

@inproceedings{zhai2024sgbot,
  author={Zhai, Guangyao and Cai, Xiaoni and Huang, Dianye and Di, Yan and Manhardt, Fabian and Tombari, Federico and Navab, Nassir and Busam, Benjamin},
  booktitle={Proc. IEEE Int. Conf. Robot. Autom. (ICRA)},
  title={{SG-Bot}: Object rearrangement via coarse-to-fine robotic imagination on scene graphs},
  year={2024},
  pages={4303--4310},
}

@misc{gao2024holistic,
  title={Toward Holistic Planning and Control Optimization for Dual-Arm Rearrangement}, 
  author={Kai Gao and Zihe Ye and Duo Zhang and Baichuan Huang and Jingjin Yu},
  note={arXiv:2404.06758},
  year={2024},
}

@inproceedings{gao2025orla,
  author={Gao, Kai and {Zhaxizhuoma} and Ding, Yan and Zhang, Shiqi and Yu, Jingjin},
  booktitle={Proc. IEEE Int. Conf. Robot. Autom. (ICRA)}, 
  title={{{ORLA}*}: Mobile Manipulator-Based Object Rearrangement with Lazy {{A}*}}, 
  year={2025},
  pages={13211--13217}
}

@inproceedings{hu2025strap,
  author={Hu, Jiaming and Szczekulski, Jan and Peddabomma, Sudhansh and Christensen, Henrik I.},
  booktitle={Proc. IEEE Int. Conf. Robot. Autom. (ICRA)}, 
  title={Planning for Tabletop Object Rearrangement}, 
  year={2025},
  pages={11889--11895}
}

@inproceedings{ren2024msmcts,
  author={Ren, Hanwen and Qureshi, Ahmed H.},
  booktitle={Proc. IEEE/RSJ Int. Conf. Intell. Robots Syst. (IROS)}, 
  title={Multi-Stage {Monte Carlo} Tree Search for Non-Monotone Object Rearrangement Planning in Narrow Confined Environments}, 
  year={2024},
  pages={12078--12085},
}

@article{chen2024osamipstack,
  author={Chen, Lawrence Yunliang and Huang, Huang and Goldberg, Ken},
  title={Optimal Arrangement and Rearrangement of Objects on Shelves to Minimize Robot Retrieval Cost}, 
  journal={IEEE Trans. Autom. Sci. Eng.}, 
  volume={21},
  number={3},
  pages={2184--2198},
  year={2024},
}

@article{chen2026testtube,
  author = {Chen, Hao and Tang, Yu and Wan, Weiwei and Matsushita, Masaki and Takahashi, Jun and Kotaka, Takeyuki and Harada, Kensuke},
  title ={Robotic test tube rearrangement using combined reinforcement learning and motion planning in a closed loop},
  journal = {Int. J. Robot. Res.},
  year = {2026},
  doi = {10.1177/02783649261430363},
}

@misc{batra2020embodied,
  title={Rearrangement: A Challenge for Embodied {AI}}, 
  author={Dhruv Batra and Angel X. Chang and Sonia Chernova and Andrew J. Davison and Jia Deng and Vladlen Koltun and Sergey Levine and Jitendra Malik and Igor Mordatch and Roozbeh Mottaghi and Manolis Savva and Hao Su},
  note={arXiv:2011.01975},
  year={2020},
}

@inproceedings{mirakhor2024multiroom,
  author = {Mirakhor, Karan and Ghosh, Sourav and Das, Dipanjan and Bhowmick, Brojeshwar},
  booktitle = {Proc. AAAI Conf. Artif. Intell.},
  title = {Task Planning for Object Rearrangement in Multi-Room Environments},
  volume = {38},
  number = {9},
  pages = {10350--10357},
  year = {2024},
}

@article{wang2020scenemover,
  author = {Wang, Hanqing and Liang, Wei and Yu, Lap-Fai},
  title = {{Scene Mover}: automatic move planning for scene arrangement by deep reinforcement learning},
  journal = {ACM Trans. Graph.},
  volume = {39},
  number = {6},
  pages = {1--15},
  year = {2020}
}

@inproceedings{kang2023retrieval,
  author={Kang, Minjae and Kim, Junseok and Kee, Hogun and Oh, Songhwai},
  booktitle={Proc. IEEE/RSJ Int. Conf. Intell. Robots Syst. (IROS)}, 
  title={Object Rearrangement Planning for Target Retrieval in a Confined Space with Lateral View}, 
  year={2023},
  pages={2004--2009},
}

@inproceedings{cheong2020relocate,
  author={Cheong, Sang Hun and Cho, Brian Y. and Lee, Jinhwi and Kim, ChangHwan and Nam, Changjoo},
  booktitle={Proc. IEEE Int. Conf. Robot. Autom. (ICRA)}, 
  title={Where to relocate?: Object rearrangement inside cluttered and confined environments for robotic manipulation}, 
  year={2020},
  pages={7791--7797},
}

@article{lee2022retrieval,
  author={Lee, Jinhwi and Rakhman, Ulzhalgas and Nam, Changjoo and Kang, Seunghyun and Park, Jonghyeon and Kim, ChangHwan},
  title={High dimensional object rearrangement for a robot manipulation in highly dense configurations},
  journal={Intell. Serv. Robot.},
  volume={15},
  number={5},
  pages={649--660},
  year={2022},
}

@inproceedings{ren2024retrieval,
  author={Ren, Hanwen and Qureshi, Ahmed H.},
  booktitle={Proc. IEEE Int. Conf. Robot. Autom. (ICRA)}, 
  title={Neural Rearrangement Planning for Object Retrieval from Confined Spaces Perceivable by Robot's In-hand {RGB-D} Sensor}, 
  year={2024},
  pages={15388--15394},
}

@inproceedings{ghasemi2024scene,
  author={Ghasemi, Seraj and Hosseini, Hamed and Koosheshi, MohammadHossein and Tale Masouleh, Mehdi and Kalhor, Ahmad},
  booktitle={Proc. 32nd Int. Conf. Elect. Eng. (ICEE)}, 
  title={Scene Understanding in Pick-and-Place Tasks: Analyzing Transformations Between Initial and Final Scenes}, 
  year={2024},
  pages={1--7}
}

@misc{yolov5,
  author = {Jocher, Glenn},
  title = {{YOLOv5} by Ultralytics},
  version = {7.0},
  month = {May},
  year = {2020},
  license = {AGPL-3.0},
  url = {https://github.com/ultralytics/yolov5},
  doi = {10.5281/zenodo.3908559},
  note = {Software}
}

@misc{gemini,
  title={Gemini: A family of highly capable multimodal models},
  author={{Gemini Team} and others},
  note={arXiv:2312.11805},
  year={2023},
}

@inproceedings{viola2001sat,
  author={Viola, P. and Jones, M.},
  booktitle={Proc. IEEE Comput. Soc. Conf. Comput. Vis. Pattern Recognit. (CVPR)}, 
  title={Rapid object detection using a boosted cascade of simple features}, 
  volume={1},
  year={2001},
  pages = {I-511--I-518}
}

@misc{coumans2021pybullet,
  title = {{PyBullet}, a {Python} module for physics simulation for games, robotics and machine learning},
  author = {Coumans, Erwin and Bai, Yunfei},
  year = {2021},
  url = {http://pybullet.org}
}

@article{gaz2019panda,
  author={Gaz, Claudio and Cognetti, Marco and Oliva, Alexander and Robuffo Giordano, Paolo and De Luca, Alessandro},
  title={Dynamic Identification of the {Franka Emika Panda} Robot With Retrieval of Feasible Parameters Using Penalty-Based Optimization}, 
  journal={IEEE Robot. Autom. Lett.}, 
  volume={4},
  number={4},
  pages={4147--4154},
  year={2019},
}

@article{dastjerdi2020delta,
  author = {Amir {Hashemi Dastjerdi} and Mohammad {Morad Sheikhi} and Mehdi {Tale Masouleh}},
  title = {A complete analytical solution for the dimensional synthesis of {3-DOF} {Delta} parallel robot for a prescribed workspace},
  journal = {Mech. Mach. Theory},
  volume = {153},
  pages = {103991},
  year = {2020},
}

\end{document}